\documentclass[10pt,twocolumn,letterpaper]{article}

\usepackage{cvpr}
\usepackage{times}
\usepackage{epsfig}
\usepackage{graphicx}
\usepackage{amsmath}
\usepackage{amssymb}
\usepackage{multirow}
\usepackage{rotating}
\usepackage{enumitem}
\usepackage{moresize}

\usepackage[dvipsnames]{xcolor}
\usepackage{colortbl}
\definecolor{todo_color}{rgb}{.8,.3,.1}

\usepackage{pifont}

\newcommand{\RR}{\mathbb{R}}
\newcommand{\LL}{\mathcal{L}}
\newcommand{\EE}{\mathop{\mathbb{E}}}

\usepackage[pagebackref=true,breaklinks=true,letterpaper=true,colorlinks,bookmarks=false]{hyperref}

\cvprfinalcopy 

\def\cvprPaperID{764} 

\begin{document}

\title{Image Generation from Scene Graphs}

\author{
  Justin Johnson\textsuperscript{1,2}\footnotemark \hspace{2.5pc}
  Agrim Gupta\textsuperscript{1} \hspace{2.5pc}
  Li Fei-Fei\textsuperscript{1,2} \\*[4pt]
  \textsuperscript{1}Stanford University \hspace{2.5pc}
  \textsuperscript{2}Google Cloud AI
}

\maketitle
\renewcommand*{\thefootnote}{\fnsymbol{footnote}}
\setcounter{footnote}{1}
\footnotetext{Work done during an internship at Google Cloud AI.}
\renewcommand*{\thefootnote}{\arabic{footnote}}
\setcounter{footnote}{0}

\begin{abstract}
  To truly understand the visual world our models should be able not only to recognize images but also generate them. To this end, there has been exciting recent progress on generating images from natural language descriptions. These methods give stunning results on limited domains such as descriptions of birds or flowers, but struggle to faithfully reproduce complex sentences with many objects and relationships. To overcome this limitation we propose a method for generating images from scene graphs, enabling explicitly reasoning about objects and their relationships. Our model uses graph convolution to process input graphs, computes a scene layout by predicting bounding boxes and segmentation masks for objects, and converts the layout to an image with a cascaded refinement network. The network is trained adversarially against a pair of discriminators to ensure realistic outputs. We validate our approach on Visual Genome and COCO-Stuff, where qualitative results, ablations, and user studies demonstrate our method's ability to generate complex images with multiple objects.

\end{abstract}
\vspace{-2mm}

\section{Introduction}
\noindent\emph{What I cannot create, I do not understand}

-- Richard Feynman \\*[-2mm]

The act of \emph{creation} requires a deep understanding of the thing being
created: chefs, novelists, and filmmakers must understand food, writing, and film at a much deeper level than diners, readers, or moviegoers. If our computer vision systems are to truly understand the visual world, they must be able not only
\emph{recognize} images but also to \emph{generate} them.

Aside from imparting deep visual understanding, methods for generating
realistic images can also be practically useful. In the near term, automatic image generation can aid the work of artists or graphic designers. One day, we might replace image and video search engines with algorithms that generate customized images and videos in response to the individual tastes of each user.

As a step toward these goals, there has been exciting recent progress on \emph{text to image synthesis}~\cite{reed2016learning,reed2016generative,reed2017parallel,zhang2017stackgan} by combining recurrent neural networks and Generative Adversarial Networks~\cite{goodfellow2014generative} to generate images from natural language descriptions.

\begin{figure}
  \centering
  \includegraphics[width=0.42\textwidth]{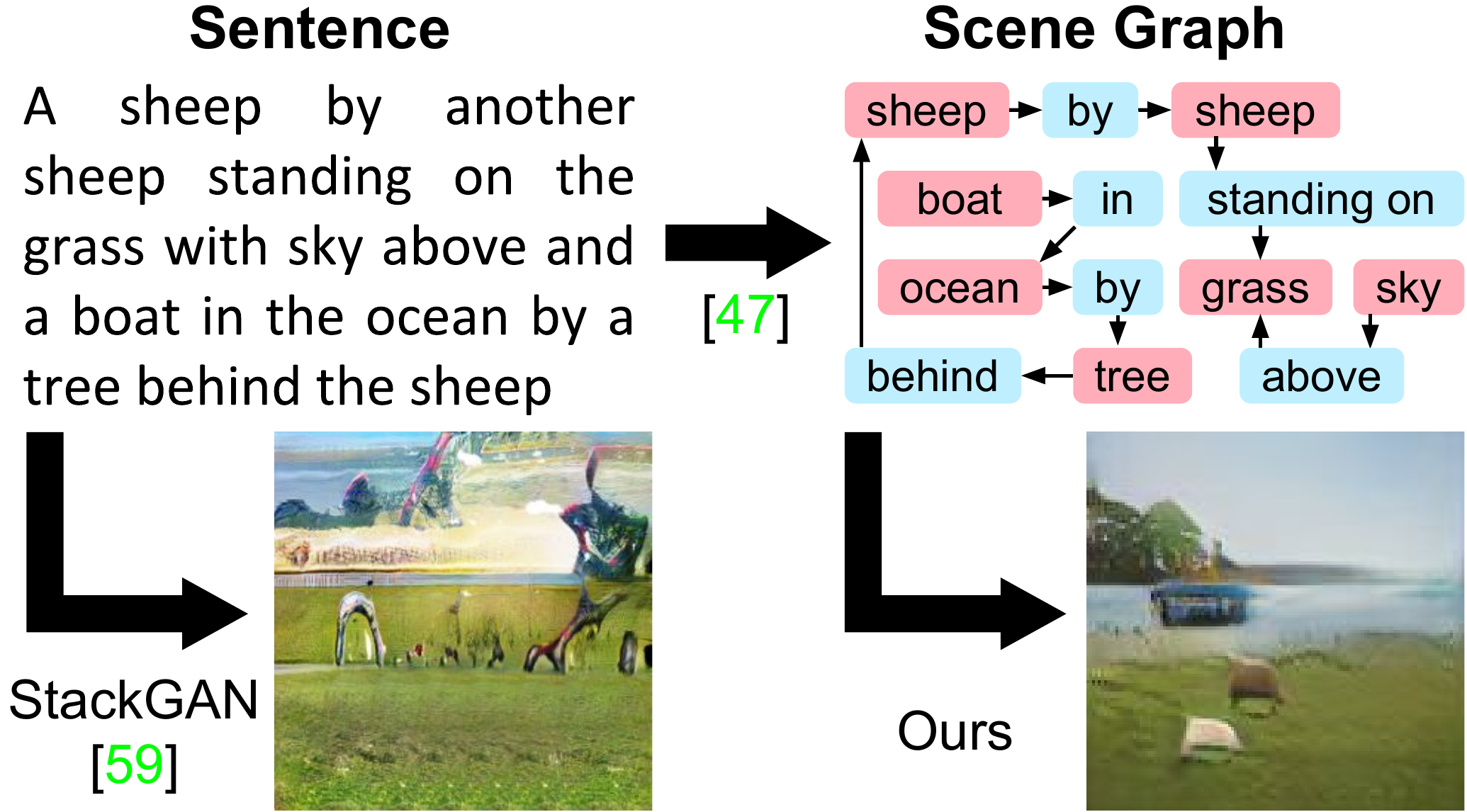}
  \caption{
    State-of-the-art methods for generating images from sentences, such as
    StackGAN~\cite{zhang2017stackgan}, struggle to faithfully depict complex
    sentences with many objects. We overcome this limitation by generating images
    from \emph{scene graphs}, allowing our method to reason explicitly about
    objects and their relationships.
  }
  \vspace{-2mm}
  \label{fig:pull}
\end{figure}

These methods can give stunning results on limited domains, such as fine-grained descriptions of birds or flowers. However as shown in Figure~\ref{fig:pull}, leading methods for generating images from sentences struggle with complex sentences containing many objects.

A sentence is a linear structure, with one word following another; 
however as shown in Figure~\ref{fig:pull}, the information conveyed by a complex
sentence can often be more explicitly represented as a \emph{scene graph} of
objects and their relationships. Scene graphs are a powerful structured 
representation for both images and language; they have been used
for semantic image retrieval~\cite{johnson2015image} and for evaluating~\cite{anderson2016spice} and improving~\cite{liu2017improved} image captioning; methods have also been developed for converting sentences to scene graphs~\cite{schuster2015generating} and for predicting scene graphs from images~\cite{lu2016visual,newell2017pixels,xu2017scene,yang2017support}.

In this paper we aim to generate complex images with many objects and
relationships by conditioning our generation on scene graphs, allowing our
model to reason explicitly about objects and their relationships.

With this new task comes new challenges. We must develop a method for processing
scene graph inputs; for this we use a \emph{graph convolution network} which passes information along graph edges. After processing the graph, we must bridge the gap between the symbolic graph-structured input and the two-dimensional image output; to this end we construct a \emph{scene layout} by predicting bounding boxes and segmentation masks for all objects in the graph. Having predicted a layout, we must generate an image which respects it; for this we use a \emph{cascaded refinement network} (CRN)~\cite{chen2017photographic} which processes the layout at increasing spatial scales. Finally, we must ensure that our generated images are realistic and contain recognizable objects; we therefore train adversarially against a pair of \emph{discriminator} networks operating on image patches and generated objects. All components of the model are learned jointly in an end-to-end manner.

We experiment on two datasets: Visual Genome~\cite{krishna2017visual},
which provides human annotated scene graphs, and 
COCO-Stuff~\cite{caesar2016coco} where we construct synthetic scene graphs from
ground-truth object positions. On both datasets we show qualitative results
demonstrating our method's ability to generate complex images which respect the
objects and relationships of the input scene graph, and perform comprehensive
ablations to validate each component of our model.

Automated evaluation of generative images models is a challenging
problem unto itself~\cite{theis2016note}, so we also evaluate our results with
two user studies on Amazon Mechanical Turk. Compared to StackGAN~\cite{zhang2017stackgan}, a leading system for text to image synthesis, users find that our results better match COCO captions in 68\% of trials, and contain 59\% more recognizable objects.

\section{Related Work}
\textbf{Generative Image Models} fall into three recent categories:
Generative Adversarial Networks (GANs)~\cite{goodfellow2014generative,radford2016unsupervised} jointly learn a \emph{generator} for synthesizing images and a \emph{discriminator} classifying images as real or fake;
Variational Autoencoders~\cite{kingma2014auto} 
use variational inference to jointly learn an \emph{encoder} and \emph{decoder} mapping between images and latent codes;
autoregressive approaches~\cite{oord2016pixel,oord2016conditional} model likelihoods by conditioning each pixel on all previous pixels.

\textbf{Conditional Image Synthesis}
conditions generation on additional input.
GANs can be conditioned on category labels by providing labels as an additional input to both generator and discriminator~\cite{gauthier2014conditional,mirza2014conditional} or by forcing the discriminator to predict the label~\cite{odena2017conditional}; we take the latter approach.

Reed \etal~\cite{reed2016generative} generate images from text using a GAN; Zhang \etal~\cite{zhang2017stackgan} extend this approach to higher resolutions using multistage generation. Related to our approach, Reed \etal generate images conditioned on sentences and keypoints using both GANs~\cite{reed2016learning} and multiscale autoregressive models~\cite{reed2017parallel}; in addition to generating images they also predict locations of unobserved keypoints using a separate generator and discriminator operating on keypoint locations.

Chen and Koltun~\cite{chen2017photographic} generate high-resolution images of street scenes from ground-truth semantic segmentation using a cascaded refinement network (CRN) trained with a perceptual feature reconstruction loss~\cite{gatys2016image,johnson2016perceptual}; we use their CRN architecture to generate images from scene layouts.

Related to our layout prediction, Chang \etal have investigated text to 3D scene generation~\cite{chang2015text,chang2014learning};
other approaches to image synthesis include stochastic grammars~\cite{jiang2017configurable}, probabalistic programming~\cite{kulkarni2015picture}, inverse graphics~\cite{kulkarni2015deep}, neural de-rendering~\cite{wu2017neural}, and generative ConvNets~\cite{xie2016theory}.

\textbf{Scene Graphs} represent scenes as directed graphs, where nodes are
objects and edges give relationships between objects. Scene graphs have been used for image retrieval~\cite{johnson2015image} and to evaluate image captioning~\cite{anderson2016spice}; some work converts sentences to scene graphs~\cite{schuster2015generating} or predicts grounded scene graphs for images~\cite{lu2016visual,newell2017pixels,xu2017scene,yang2017support}. Most work on scene graphs uses the Visual Genome dataset~\cite{krishna2017visual}, which provides human-annotated scene graphs.

\textbf{Deep Learning on Graphs.} Some methods learn embeddings for graph nodes given a single large graph~\cite{perozzi2014deepwalk,tang2015line,grover2016node2vec} similar to word2vec~\cite{mikolov2013distributed} which learns embeddings for words given a text corpus. These differ from our approach, since we must process a new graph on each forward pass.

More closely related to our work are Graph Neural Networks (GNNs)~\cite{goller1996learning,gori2005new,scarselli2009graph} which generalize recursive neural networks~\cite{frasconi1998general,sperduti1997supervised,socher2011parsing} to operate on arbitrary graphs. GNNs and related models have been applied to molecular property prediction~\cite{duvenaud2015convolutional}, program verification~\cite{li2016gated}, modeling human motion~\cite{jain2016structural}, and premise selection for theorem proving~\cite{wang2017premise}. Some methods operate on graphs in the spectral domain~\cite{bruna2014spectral,henaff2015deep,kipf2017semi} though we do not take this approach.

\section{Method}
Our goal is to develop a model which takes as input a \emph{scene graph}
describing objects and their relationships, and which generates a realistic
image corresponding to the graph. The primary challenges are threefold: first,
we must develop a method for processing the graph-structured input; second, we
must ensure that the generated images respect the objects and relationships
specified by the graph; third, we must ensure that the synthesized images are realistic.

\begin{figure*}
  \centering
  \includegraphics[width=0.93\textwidth]{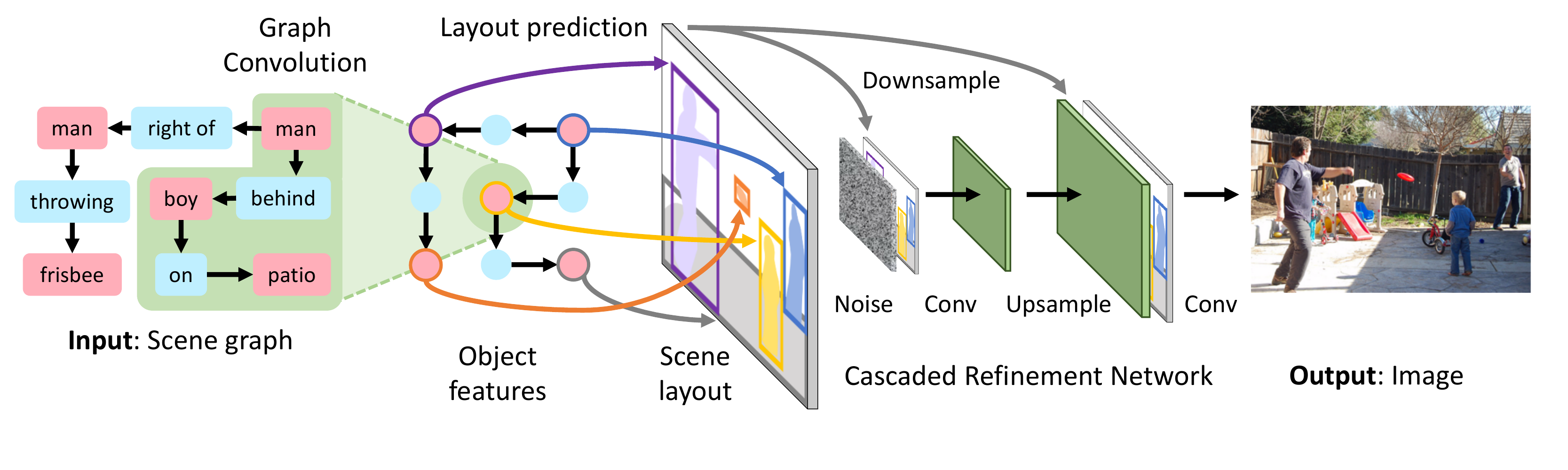}
  \vspace{-4mm}
  \caption{
    Overview of our image generation network $f$ for generating images from
    scene graphs.
    The input to the model is a \emph{scene graph} specifying objects and relationships;
    it is processed with a \emph{graph convolution network} (Figure~\ref{fig:gconv}) which passes information along
    edges to compute embedding vectors for all objects. These vectors are used to predict
    bounding boxes and segmentation masks for objects, which are combined to form a
    \emph{scene layout} (Figure~\ref{fig:layout}). The layout is converted to an image using a \emph{cascaded refinement network}
    (CRN)~\cite{chen2017photographic}. The model is trained adversarially against a pair of
    \emph{discriminator networks}.
    During training the model observes ground-truth object bounding boxes and (optionally) segmentation masks, but these are predicted by the model at test-time.
  }
  \vspace{-2mm}
  \label{fig:generator}
\end{figure*}

We convert scene graphs to images with an \emph{image generation network} $f$,
shown in Figure~\ref{fig:generator}, which inputs a scene graph $G$ and noise
$z$ and outputs an image $\hat I = f(G, z)$.

\begin{figure}[t]
  \centering
  \includegraphics[width=0.34\textwidth]{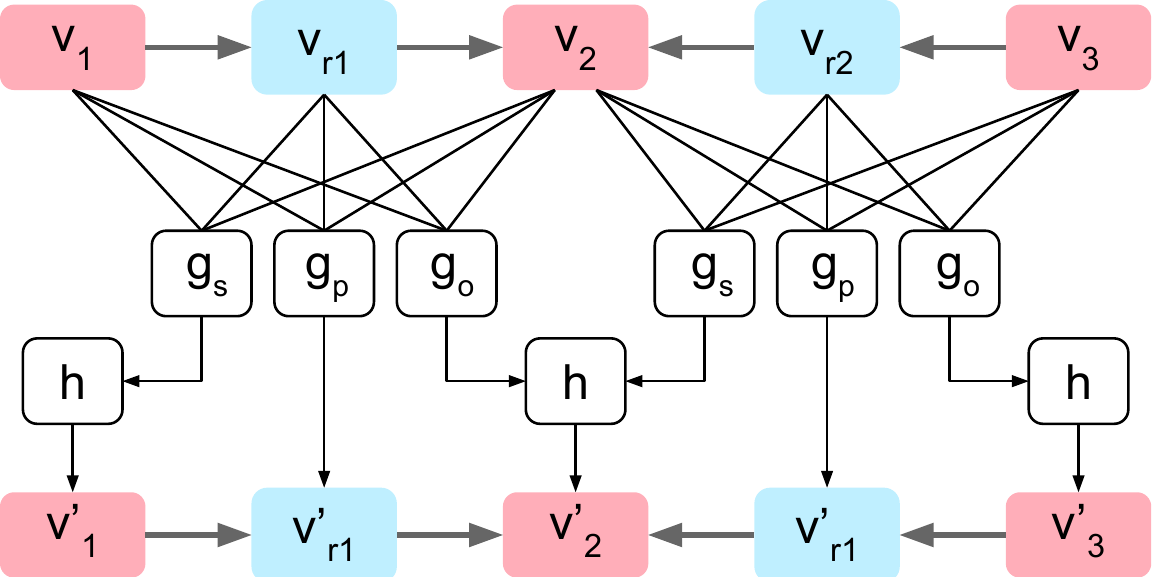}
  \caption{
    Computational graph illustrating a single graph convolution layer.
    The graph consists of three objects $o_1$, $o_2$, and $o_3$ and two
    edges $(o_1, r_1, o_2)$ and $(o_3, r_2, o_2)$. Along each edge, the three
    input vectors are passed to functions $g_s$, $g_p$, and $g_o$; $g_p$ directly
    computes the output vector for the edge, while $g_s$ and $g_o$ compute
    \emph{candidate vectors} which are fed to a symmetric pooling function $h$
    to compute output vectors for objects.
  }
  \vspace{-2mm}
  \label{fig:gconv}
\end{figure}

The scene graph $G$ is processed by a \emph{graph convolution network} which 
gives embedding vectors for each object; as shown in Figures~\ref{fig:generator}
and \ref{fig:gconv},
each layer of graph convolution mixes information along edges of the graph.

We respect the objects and relationships from $G$ by using the object
embedding vectors from the graph convolution network to predict bounding boxes
and segmentation masks for each object; these are combined to form a \emph{scene layout},
shown in the center of Figure~\ref{fig:generator}, which acts as an intermediate between
the graph and the image domains.

The output image $\hat I$ is generated from the layout using a
\emph{cascaded refinement network} (CRN)~\cite{chen2017photographic}, shown in the
right half of Figure~\ref{fig:generator}; each of its modules processes the layout
at increasing spatial scales, eventually generating the image $\hat I$.

We generate realistic images by training $f$ adversarially against a
pair of \emph{discriminator} networks $D_{img}$ and $D_{obj}$ which encourage
the image $\hat I$ to both appear realistic and to contain realistic, recognizable
objects.

Each of these components is described in more detail below; the supplementary
material describes the exact architecures used in our experiments.

\textbf{Scene Graphs.}
The input to our model is a \emph{scene graph}~\cite{johnson2015image}
describing objects and relationships between objects.
Given a set of object categories $\mathcal{C}$ and a set of relationship
categories $\mathcal{R}$, a scene graph is a tuple $(O, E)$ where
$O=\{o_1, \ldots, o_n\}$ is a set of objects with each $o_i\in\mathcal{C}$,
and $E\subseteq O\times\mathcal{R}\times O$ is a set of directed edges of
the form $(o_i, r, o_j)$ where $o_i,o_j\in O$ and $r\in\mathcal{R}$.

As a first stage of processing, we use a learned embedding layer to convert
each node and edge of the graph from a categorical label to a dense vector,
analogous to the embedding layer typically used in neural language models.

\textbf{Graph Convolution Network.}
In order to process scene graphs in an end-to-end manner, we need a neural
network module which can operate natively on graphs. To this end we use a
\emph{graph convolution network} composed of several \emph{graph convolution
layers}.

A traditional 2D convolution layer takes as input a spatial grid of feature vectors
and produces as output a new spatial grid of vectors, where each output vector is a
function of a local neighborhood of its corresponding input vector; in this way a
convolution aggregates information across local neighborhoods of the input. A single
convolution layer can operate on inputs of arbitrary shape through the use of
\emph{weight sharing} across all neighborhoods in the input.

Our graph convolution layer performs a similar function: given an input graph with
vectors of dimension $D_{in}$ at each node and edge, it computes new vectors of
dimension $D_{out}$ for each node and edge. Output vectors are a function of a
neighborhood of their corresponding inputs, so that each graph convolution layer
propagates information along edges of the graph. A graph convolution layer applies
the same function to all edges of the graph, allowing a single layer to operate on
graphs of arbitrary shape.

Concretely, given input vectors $v_i,v_r\in\RR^{D_{in}}$ for all objects $o_i\in O$ and
edges $(o_i, r, o_j)\in E$, we compute output vectors for $v_i',v_r'\in\RR^{D_{out}}$
for all nodes and edges using three functions $g_s$, $g_p$, and $g_o$, which
take as input the triple of vectors $(v_i, v_r, v_j)$ for an edge and output new vectors
for the subject $o_i$, predicate $r$, and object $o_j$ respectively.

To compute the output vectors $v_r'$ for edges we simply set $v_r'=g_p(v_i,v_r,v_j)$.
Updating object vectors is more complex, since an object may participate in
many relationships; as such the output vector $v_i'$ for an object $o_i$ should depend
on all vectors $v_j$ for objects to which $o_i$ is connected via graph edges, as well
as the vectors $v_r$ for those edges. To this end, for each edge starting at $o_i$
we use $g_s$ to compute a \emph{candidate vector}, collecting all such candidates in
the set $V_i^s$; we similarly use $g_o$ to compute a set of candidate vectors $V_i^o$
for all edges terminating at $o_i$. Concretely,
\begin{align}
  V_i^s &= \{g_s(v_i, v_r, v_j) : (o_i, r, o_j) \in E\} \\
  V_i^o &= \{g_o(v_j, v_r, v_i) : (o_j, r, o_i) \in E\}.
\end{align}
The output vector for $v_i'$ for object $o_i$ is then computed as
$v_i'=h(V_i^s\cup V_i^o)$ where $h$ is a symmetric function which pools an input set of
vectors to a single output vector. An example computational graph for a single graph
convolution layer is shown in Figure~\ref{fig:gconv}.

In our implementation, the functions $g_s$, $g_p$, and $g_o$ are implemented using
a single network which concatenates its three input vectors, feeds them to a multilayer
perceptron (MLP), and computes three output vectors using fully-connected output heads.
The pooling function $h$ averages its input vectors and feeds the result to a MLP.

\textbf{Scene Layout.}
Processing the input scene graph with a series of graph convolution layers
gives an embedding vector for each object which aggregates information across
all objects and relationships in the graph.

\begin{figure}[t]
  \centering
  \includegraphics[width=0.48\textwidth]{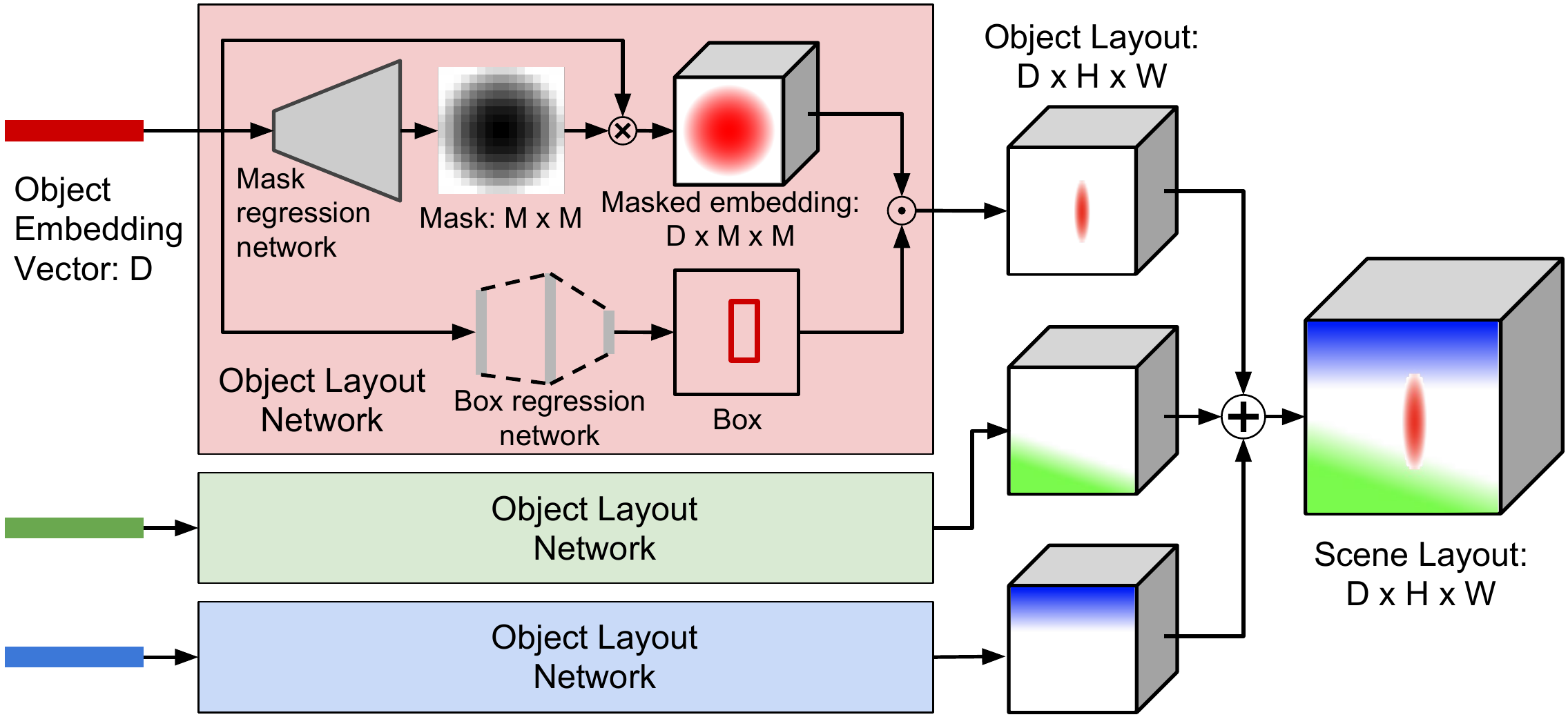}
  \caption{
    We move from the graph domain to the image domain by computing a
    \emph{scene layout}. The embedding vector for each object is passed to an
    object layout network which predicts a layout for the object;
    summing all object layouts gives the scene layout. Internally the object
    layout network predicts a soft binary segmentation mask and a bounding box
    for the object; these are combined with the embedding vector using bilinear
    interpolation to produce the object layout.
  }
  \label{fig:layout}
\end{figure}

In order to generate an image, we must move from the graph domain to the image
domain. To this end, we use the object embedding vectors to compute a
\emph{scene layout} which gives the coarse 2D structure of the image to
generate; we compute the scene layout by predicting a segmentation mask and
bounding box for each object using an \emph{object layout network}, shown in
Figure~\ref{fig:layout}.

The object layout network receives an embedding vector $v_i$ of shape $D$ for
object $o_i$ and passes
it to a \emph{mask regression network} to predict a soft binary mask $\hat m_i$ 
of shape $M\times M$ and a \emph{box regression network} to predict a bounding box
$\hat b_i=(x_0, y_0, x_1, y_1)$. The mask regression network consists of several
transpose convolutions terminating in a sigmoid nonlinearity so that
elements of the mask lies in the range $(0, 1)$; the box regression network
is a MLP. 

We multiply the embedding vector $v_i$ elementwise with the mask $\hat m_i$
to give a masked embedding of shape $D\times M\times M$ which is then warped
to the position of the bounding box using bilinear
interpolation~\cite{jaderberg2015spatial} to give an object layout.
The scene layout is then the sum of all object layouts.

During training we use ground-truth bounding boxes $b_i$ to compute the scene
layout; at test-time we instead use predicted bounding boxes $\hat b_i$.

\textbf{Cascaded Refinement Network.}
Given the scene layout, we must synthesize an image that respects the object
positions given in the layout. For this task we use a Cascaded Refinement
Network~\cite{chen2017photographic} (CRN). A CRN consists of a series of
convolutional refinement modules, with spatial resolution doubling between
modules; this allows generation to proceed in a coarse-to-fine manner.

Each module receives as input both the scene layout (downsampled to the input
resolution of the module) and the output from the previous module. These inputs
are concatenated channelwise and passed to a pair of $3\times3$ convolution
layers; the output is then upsampled using nearest-neighbor interpolation
before being passed to the next module.

The first module takes Gaussian noise $z\sim p_z$ as input, and the output from the last
module is passed to two final convolution layers to produce the output image.

\textbf{Discriminators.}
We generate realistic output images by training the image generation network $f$
adversarially against a pair of \emph{discriminator} networks $D_{img}$ and $D_{obj}$.

A discriminator $D$ attempts to classify its input $x$ as real or fake by maximizing
the objective~\cite{goodfellow2014generative}
\begin{equation}
  \LL_{GAN} = \EE_{x\sim p_{\textrm{real}}} \log D(x) + \EE_{x\sim p_{\textrm{fake}}} \log(1 - D(x))
\end{equation}
where $x\sim p_{\textrm{fake}}$ are outputs from the generation network $f$. At the same time,
$f$ attempts to generate outputs which will fool the discriminator by minimizing
$\LL_{GAN}$.\footnote{In practice, to avoid vanishing gradients $f$ typically maximizes
the surrogate objective $\EE_{x\sim p_{\textrm{fake}}} \log D(x)$ instead of minimizing
$\LL_{GAN}$~\cite{goodfellow2014generative}.}

The patch-based \emph{image discriminator} $D_{img}$ ensures that the overall
appearance of generated images is realistic; it classifies a regularly spaced,
overlapping set of image patches as real or fake, and is implemented as a fully
convolutional network, similar to the discriminator used in \cite{isola2017image}.

\newlength{\qualsize}
\setlength{\qualsize}{0.1192\textwidth}

\newlength{\qualtext}
\setlength{\qualtext}{0.11\textwidth}

\newlength{\qualtextspace}
\setlength{\qualtextspace}{0.005\textwidth}
\begin{figure*}[ht!]
  \centering
  \begin{rotate}{90}
    \textbf{Graph}
  \end{rotate}
  \hspace*{0.5mm}
  \includegraphics[width=\qualsize]{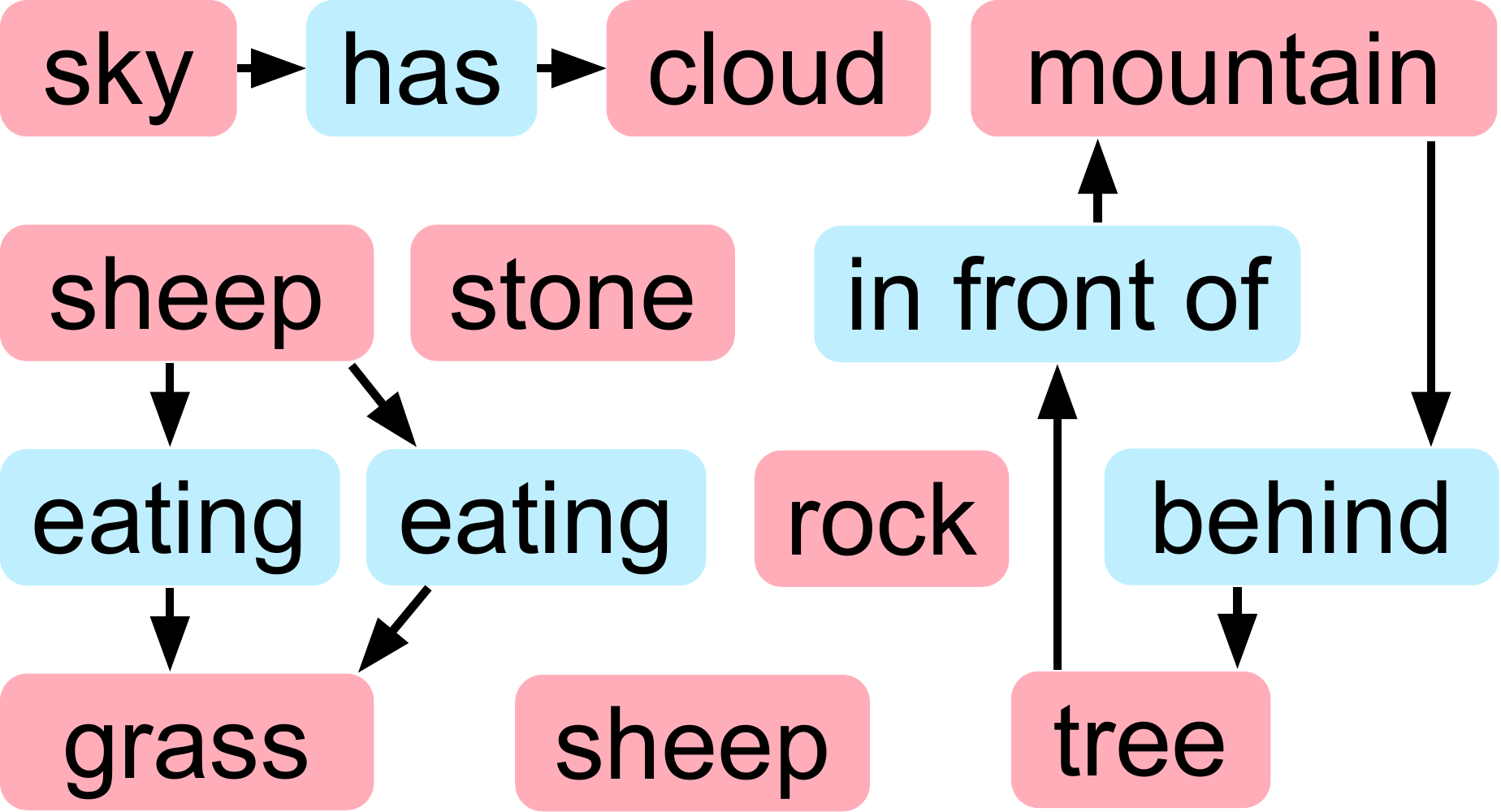}
  \includegraphics[width=\qualsize]{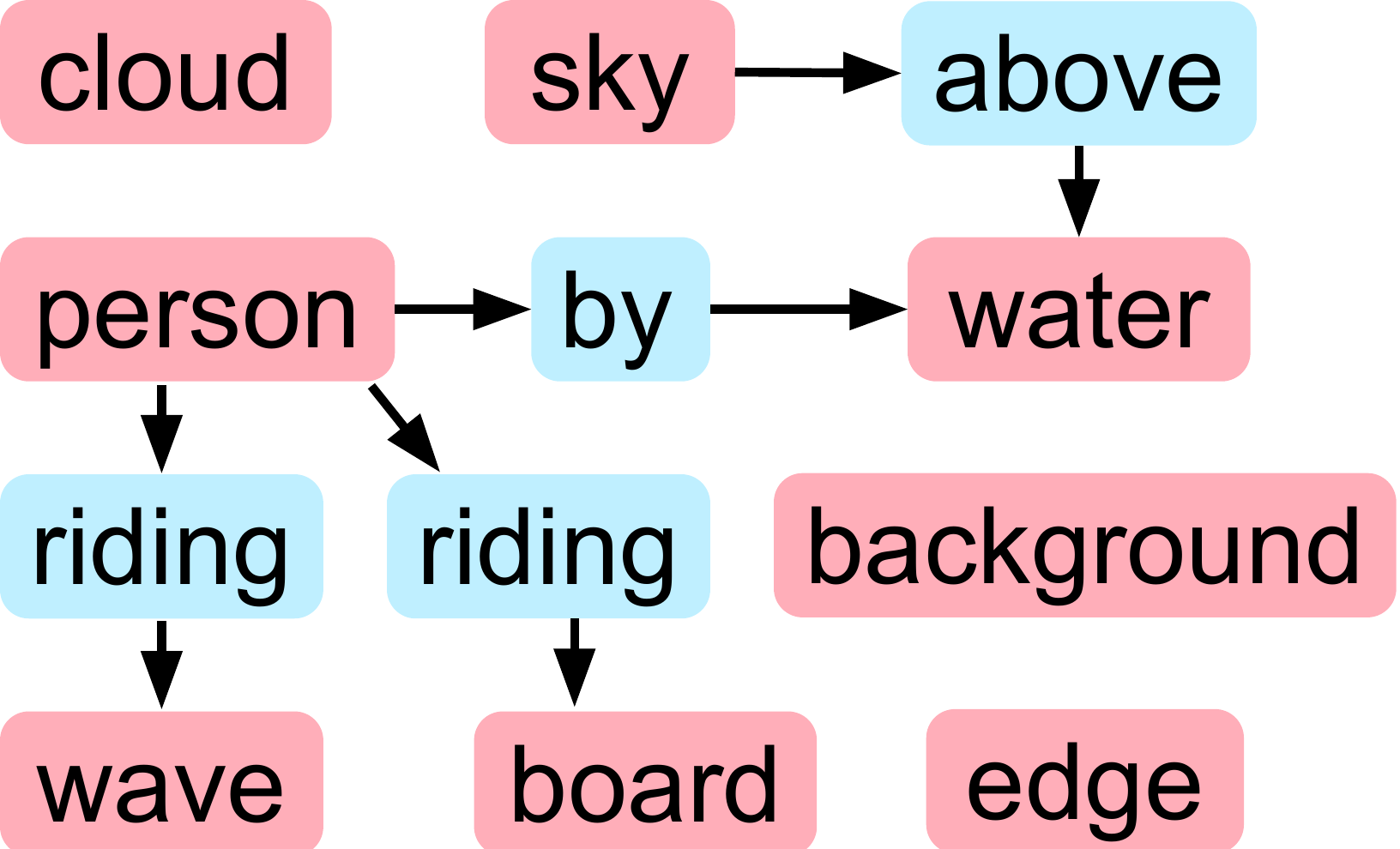}
  \includegraphics[width=\qualsize]{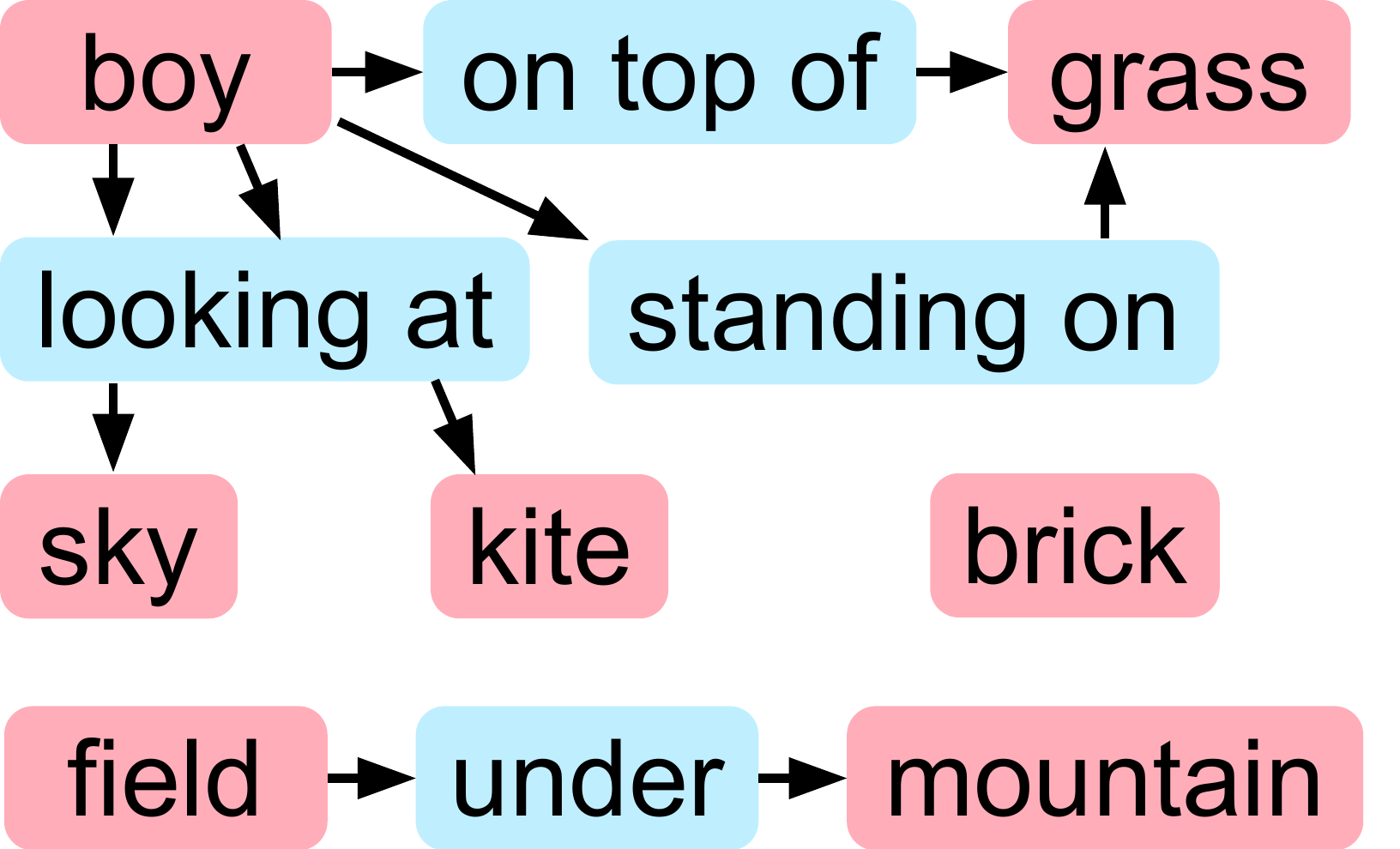}
  \includegraphics[width=\qualsize]{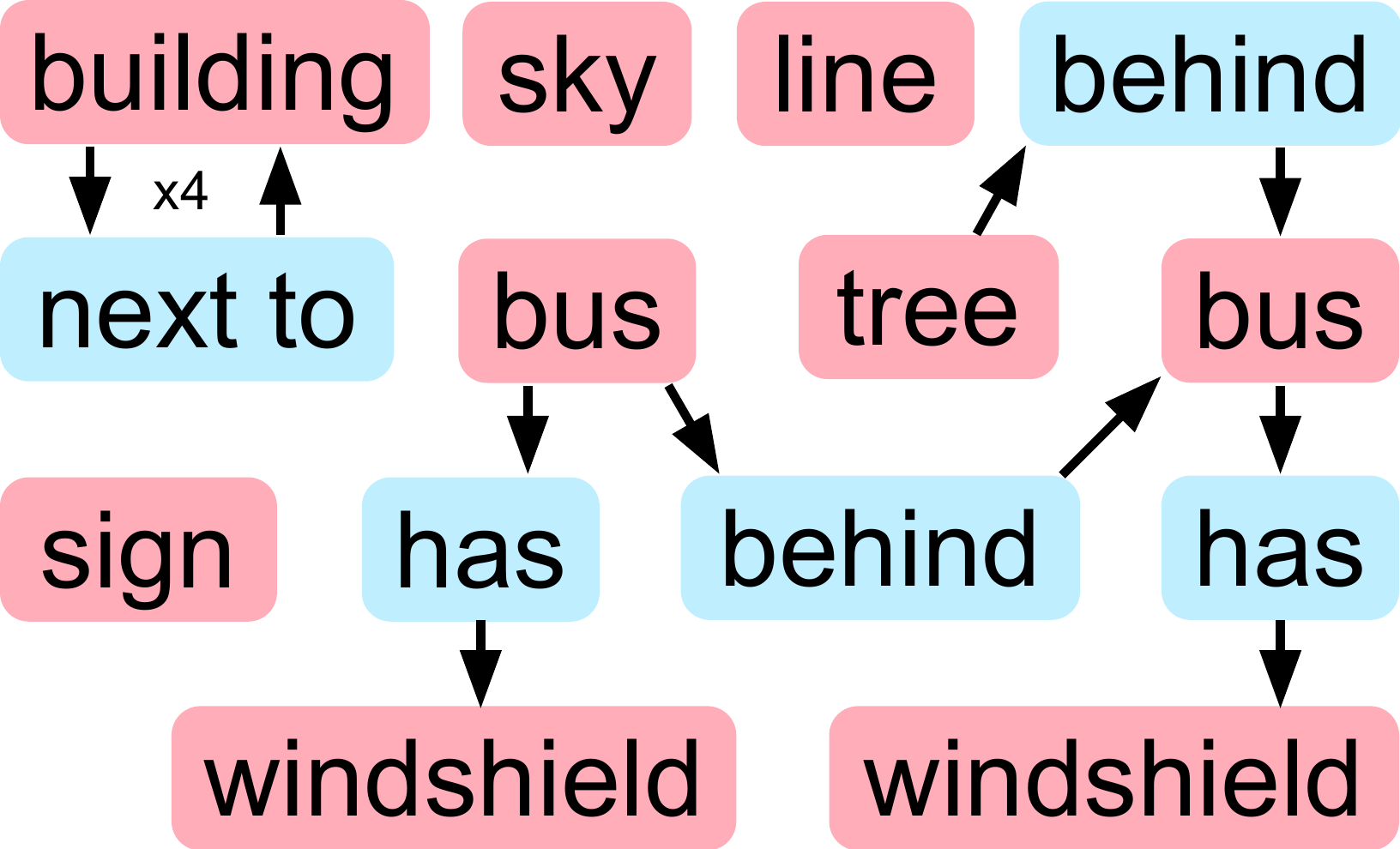}
  \includegraphics[width=\qualsize]{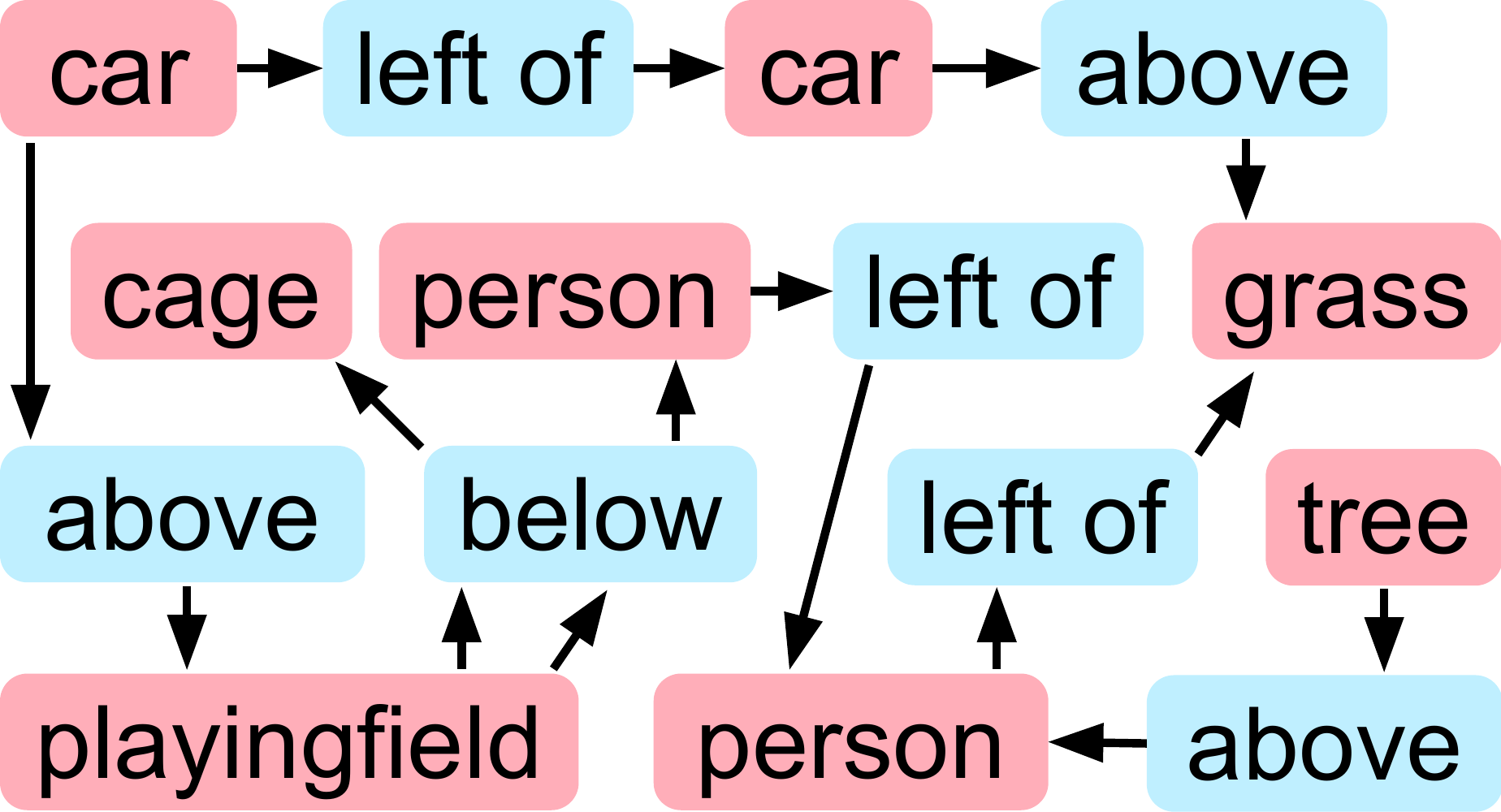}
  \includegraphics[width=\qualsize]{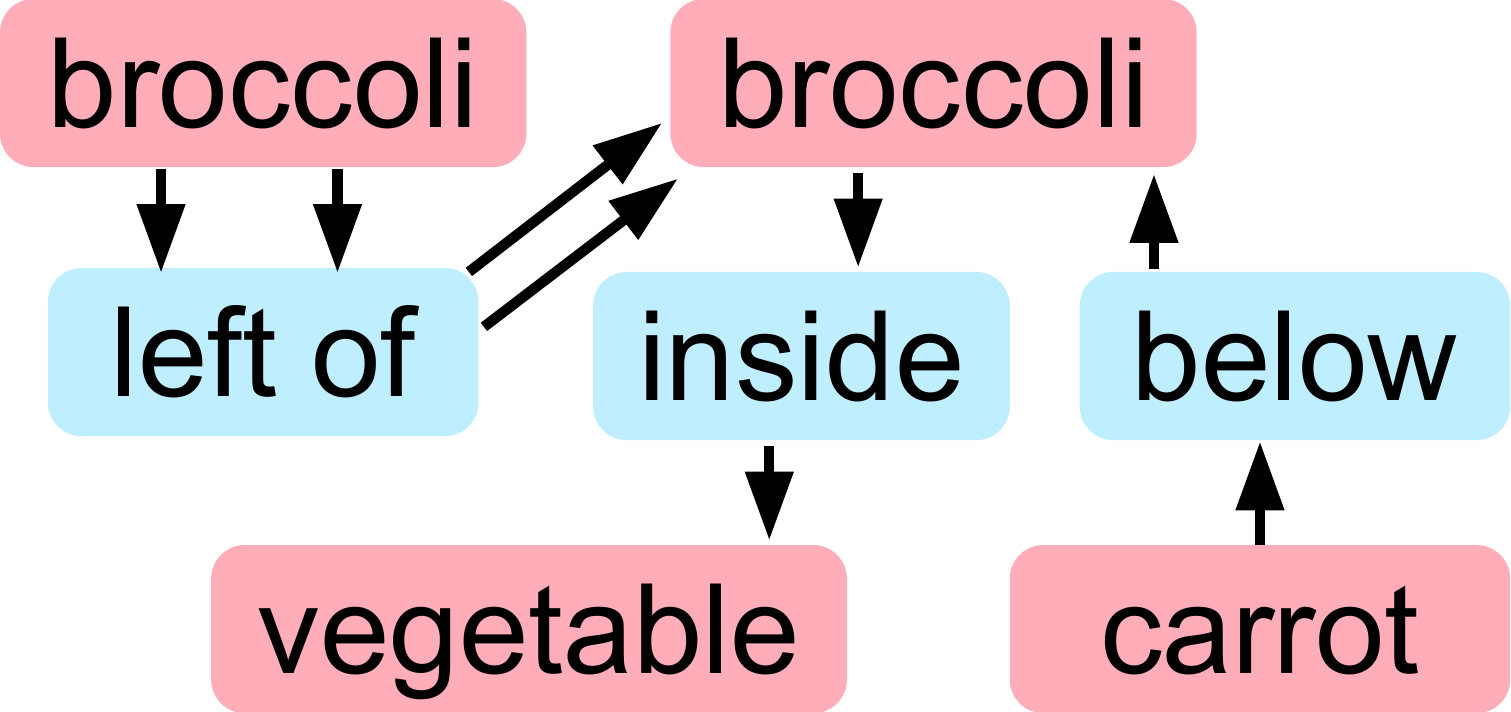}
  \includegraphics[width=\qualsize]{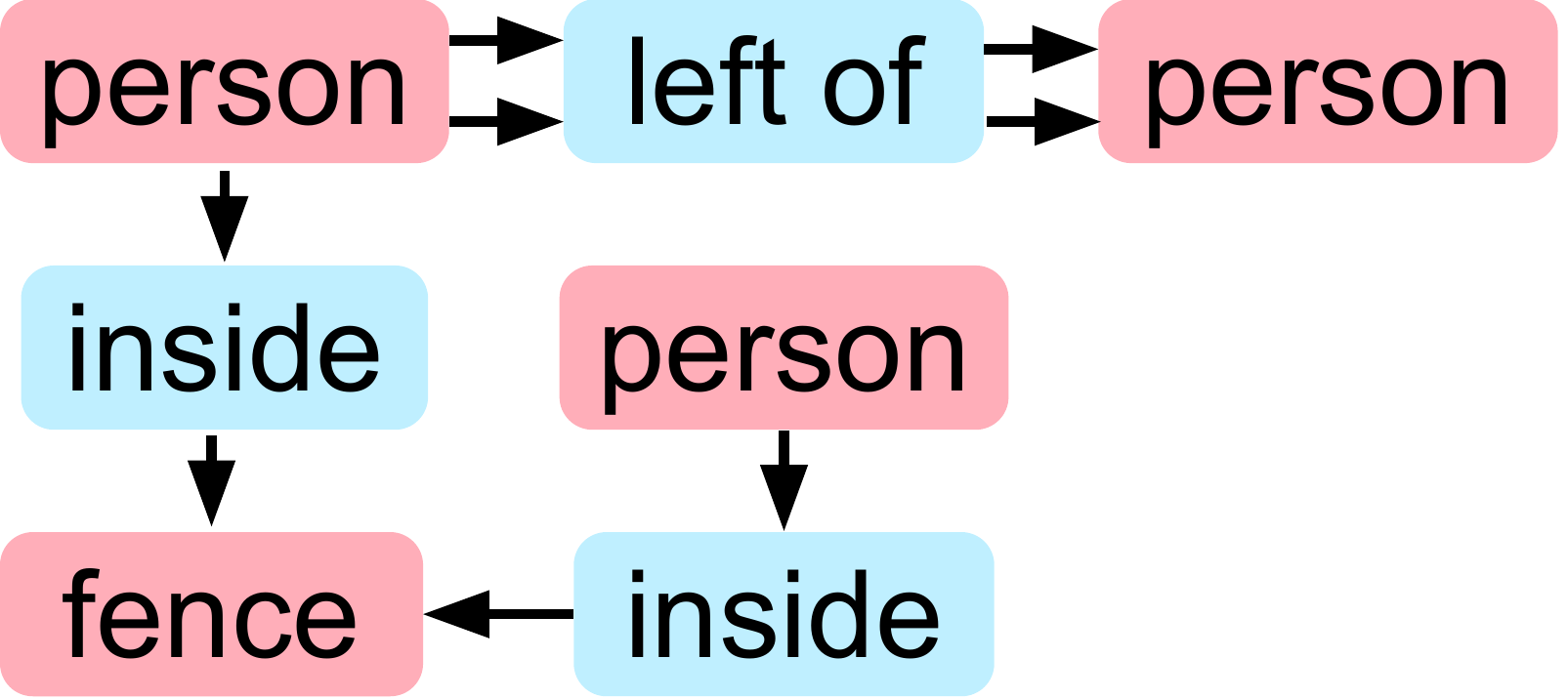}
  \includegraphics[width=\qualsize]{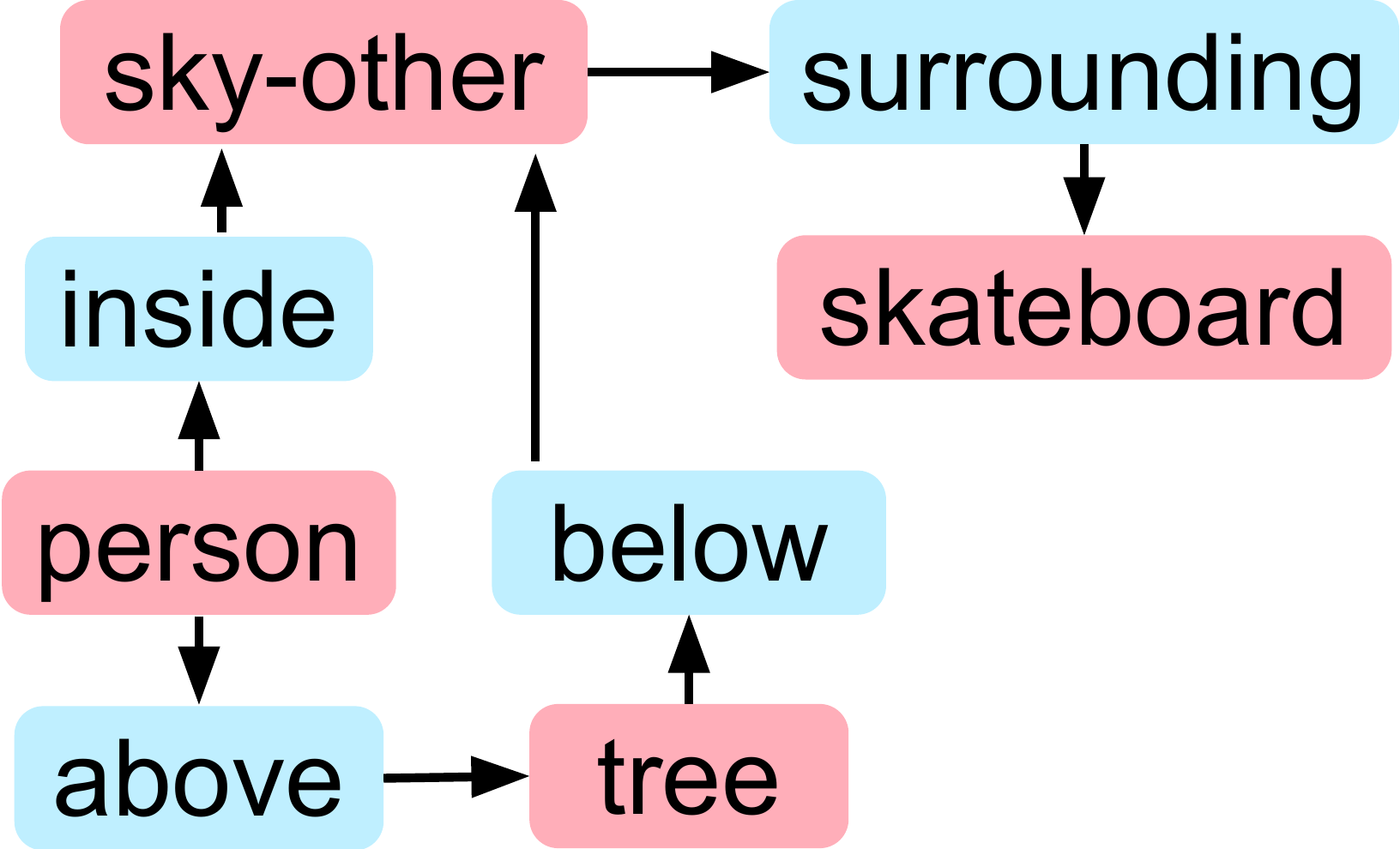} \\*[0.5mm]
  \begin{rotate}{90}
    \hspace{-2.5mm}\textbf{Text}
  \end{rotate}
  \hspace{\qualtextspace}
  \begin{minipage}{\qualtext}
    \ssmall
    Two sheep, one eating grass with a tree in front of a mountain;
    the sky has a cloud.
  \end{minipage}
  \hspace{\qualtextspace}
  \begin{minipage}{\qualtext}
    \ssmall
    A person riding a wave and a board by the water with sky above.
  \end{minipage}
  \hspace{\qualtextspace}
  \begin{minipage}{\qualtext}
    \ssmall
    A boy standing on grass looking at a kite and the sky with
    the field under a mountain
  \end{minipage}
  \hspace{\qualtextspace}
  \begin{minipage}{\qualtext}
    \ssmall
    Two busses, one behind the other and a tree behind the second;
    both busses have winshields.
  \end{minipage}
  \hspace{\qualtextspace}
  \begin{minipage}{\qualtext}
    \ssmall
    A person above a playingfield and left of another person left of grass,
    with a car left of a car above the grass.
  \end{minipage}
  \hspace{\qualtextspace}
  \begin{minipage}{\qualtext}
    \ssmall
    One broccoli left of another, which is inside vegetables
    and has a carrot below it.
  \end{minipage}
  \hspace{\qualtextspace}
  \begin{minipage}{\qualtext}
    \ssmall
    Three people with the first two inside a fence and the first
    left of the third.
  \end{minipage}
  \hspace{\qualtextspace}
  \begin{minipage}{\qualtext}
    \ssmall
    A person above the trees inside the sky, with a skateboard
    surrounded by sky.
  \end{minipage} \\
  \begin{rotate}{90}
    \hspace{4mm}
    \textbf{Layout}
  \end{rotate}
  \hspace*{0.5mm}
  \includegraphics[width=\qualsize]{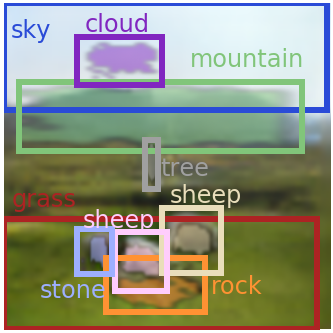}
  \includegraphics[width=\qualsize]{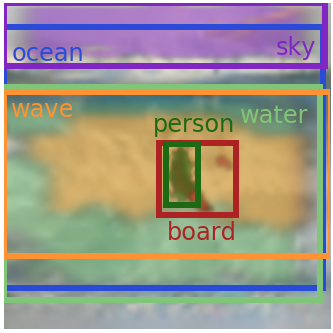}
  \includegraphics[width=\qualsize]{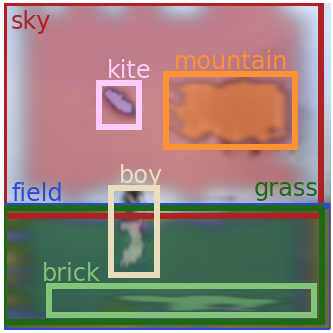}
  \includegraphics[width=\qualsize]{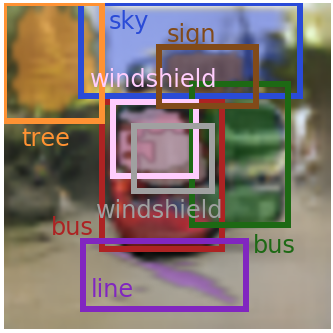}
  \includegraphics[width=\qualsize]{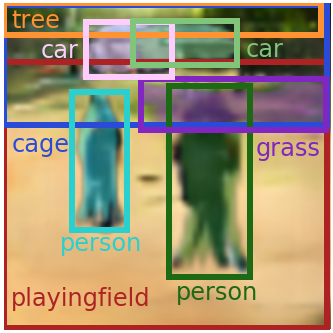}
  \includegraphics[width=\qualsize]{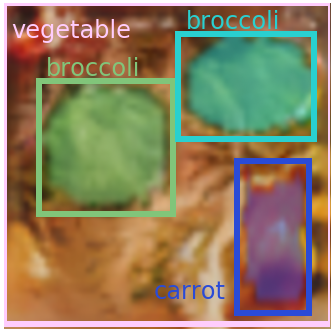}
  \includegraphics[width=\qualsize]{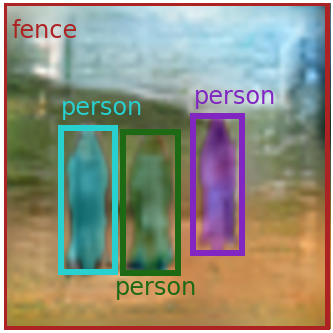}
  \includegraphics[width=\qualsize]{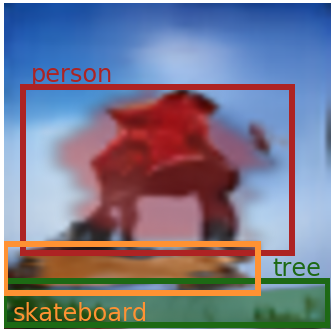} \\
  \begin{rotate}{90}
    \hspace{6mm}\textbf{Image}
  \end{rotate}
  \hspace*{0.5mm}
  \includegraphics[width=\qualsize]{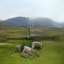}
  \includegraphics[width=\qualsize]{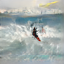}
  \includegraphics[width=\qualsize]{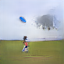}
  \includegraphics[width=\qualsize]{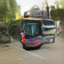}
  \includegraphics[width=\qualsize]{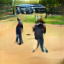}
  \includegraphics[width=\qualsize]{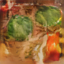}
  \includegraphics[width=\qualsize]{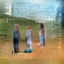}
  \includegraphics[width=\qualsize]{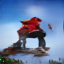} \\*[-1.5mm]
  \begin{minipage}{\qualsize} \centering\scriptsize (a) \end{minipage}
  \begin{minipage}{\qualsize} \centering\scriptsize (b) \end{minipage}
  \begin{minipage}{\qualsize} \centering\scriptsize (c) \end{minipage}
  \begin{minipage}{\qualsize} \centering\scriptsize (d) \end{minipage}
  \begin{minipage}{\qualsize} \centering\scriptsize (e) \end{minipage}
  \begin{minipage}{\qualsize} \centering\scriptsize (f) \end{minipage}
  \begin{minipage}{\qualsize} \centering\scriptsize (g) \end{minipage}
  \begin{minipage}{\qualsize} \centering\scriptsize (h) \end{minipage} \\*[1mm]
  \begin{rotate}{90}
    \hspace{1mm}
    \textbf{Graph}
  \end{rotate}
  \hspace*{0.5mm}
  \includegraphics[width=\qualsize]{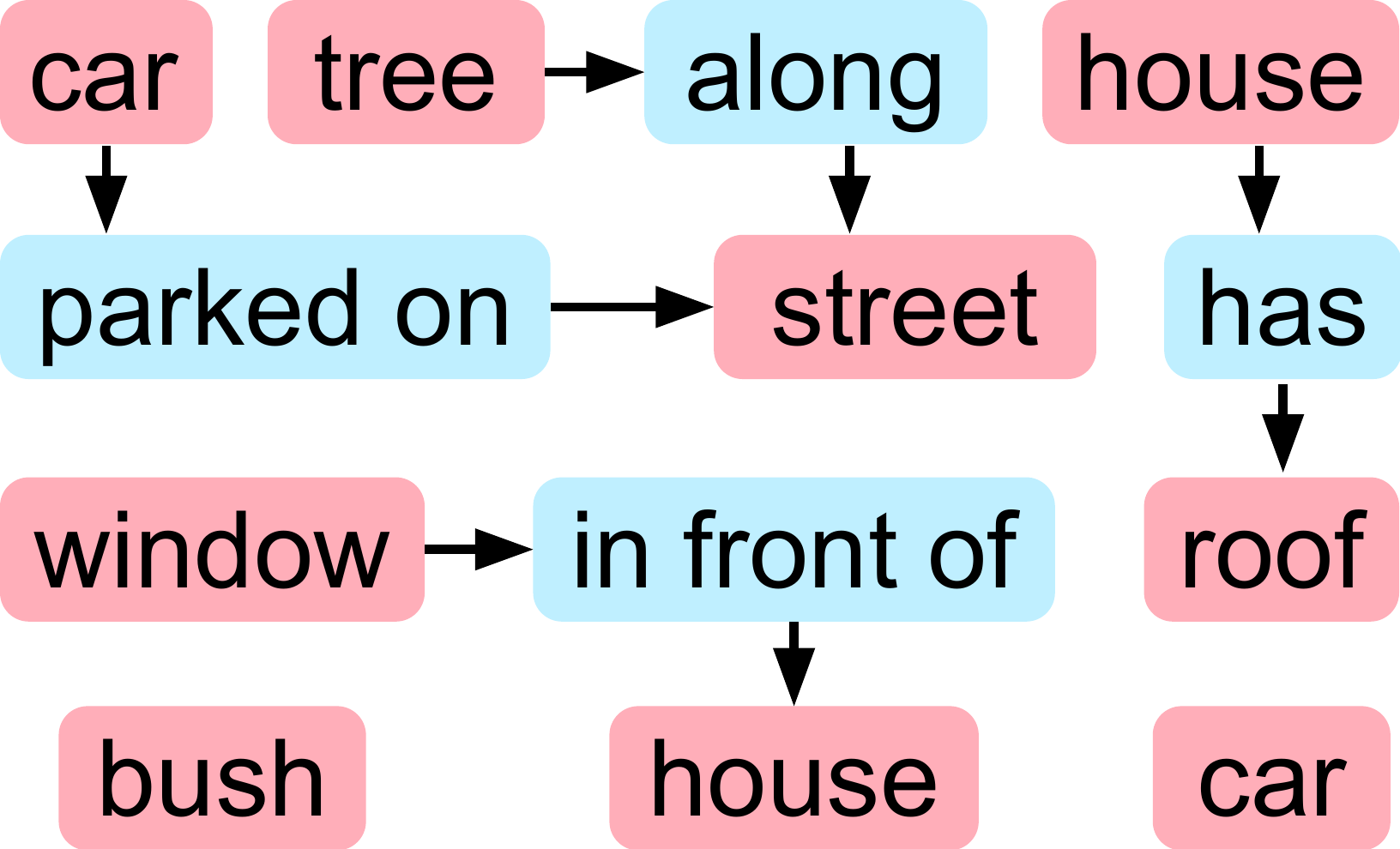}
  \includegraphics[width=\qualsize]{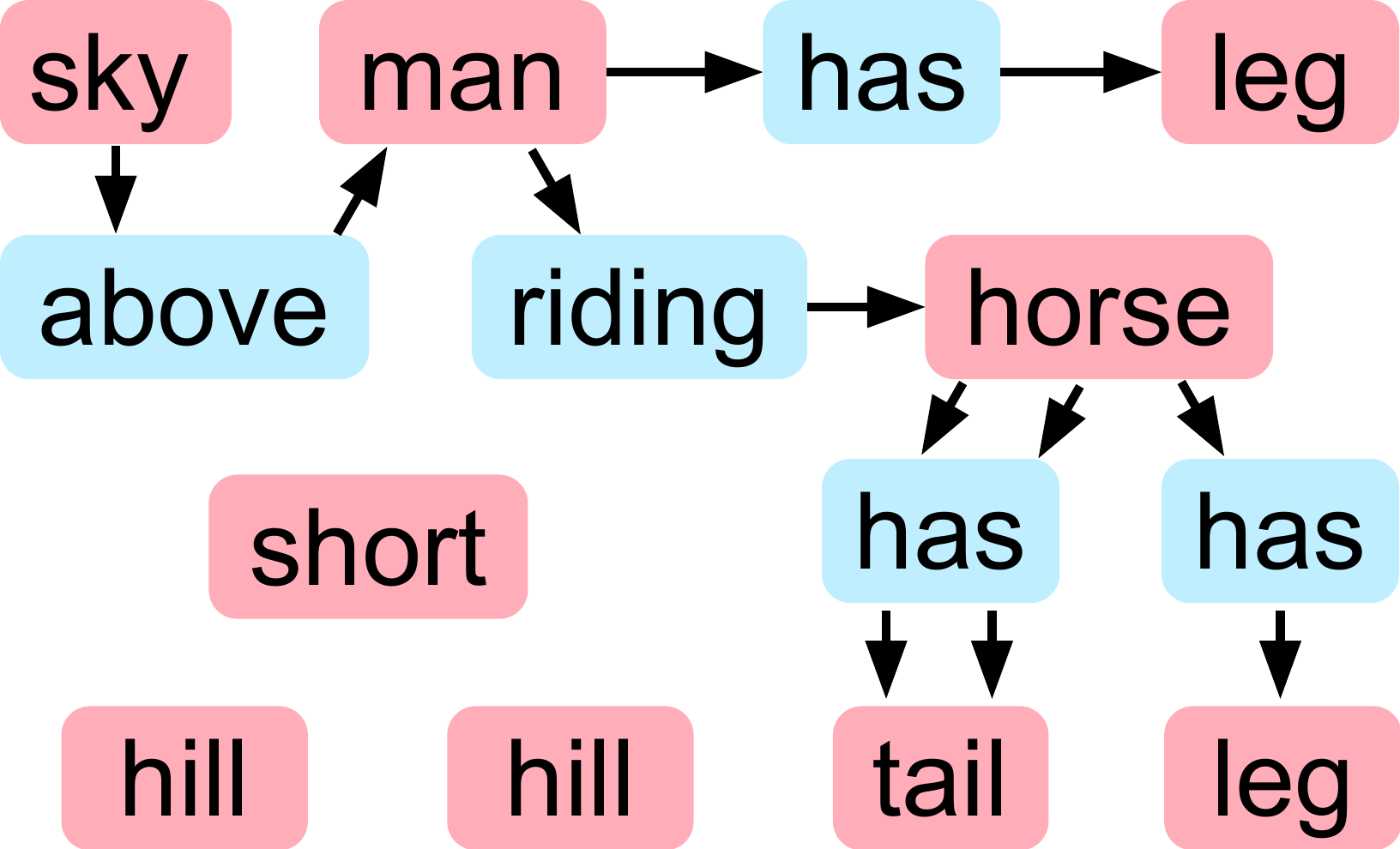}
  \includegraphics[width=\qualsize]{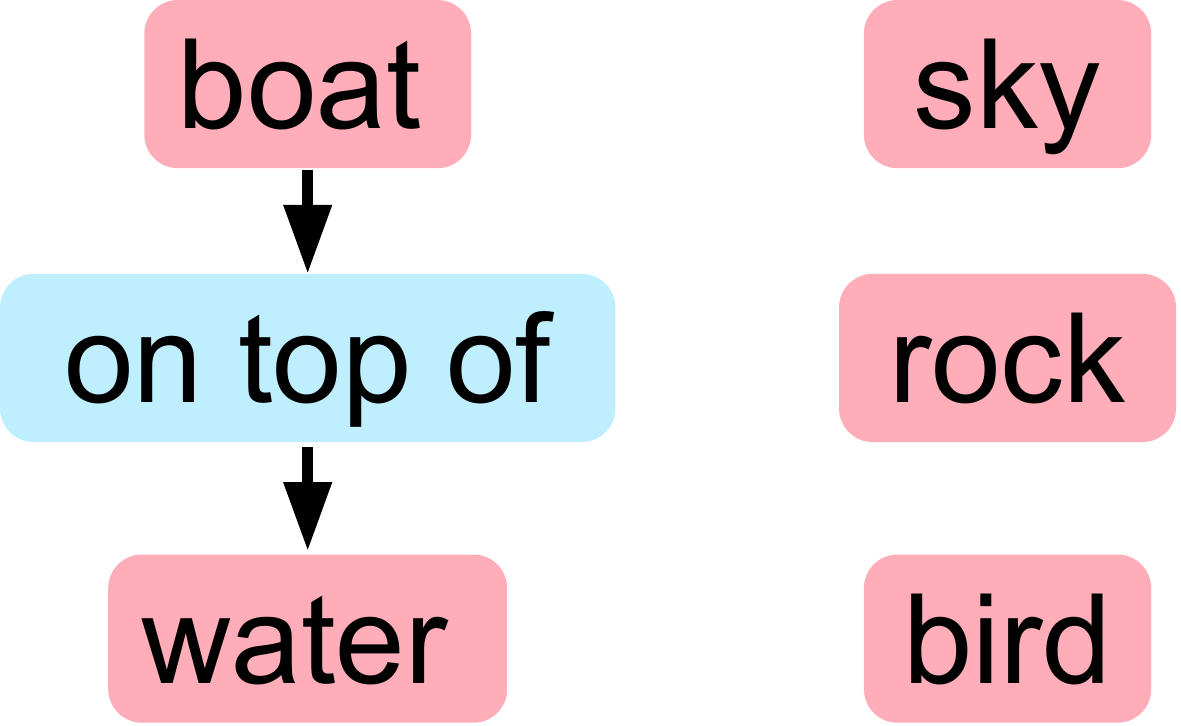}
  \includegraphics[width=\qualsize]{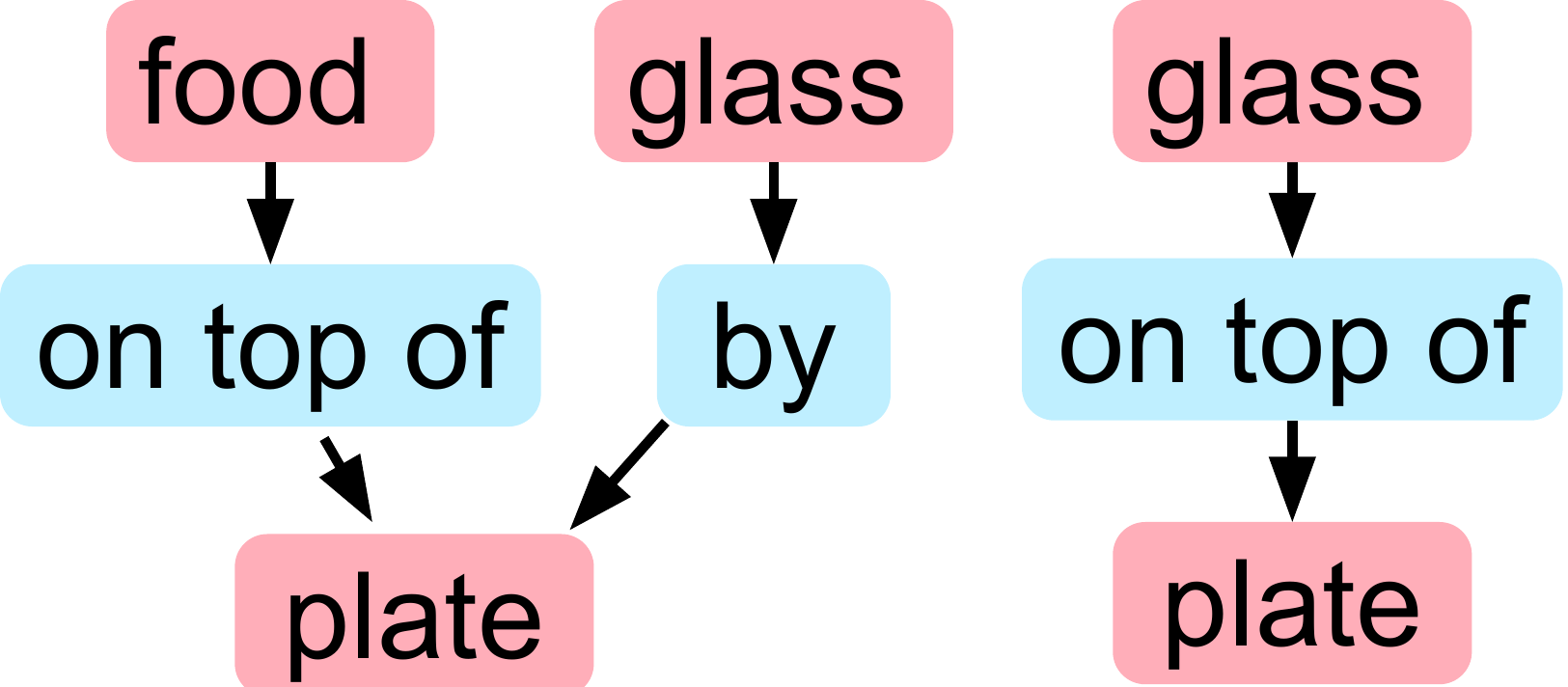}
  \includegraphics[width=\qualsize]{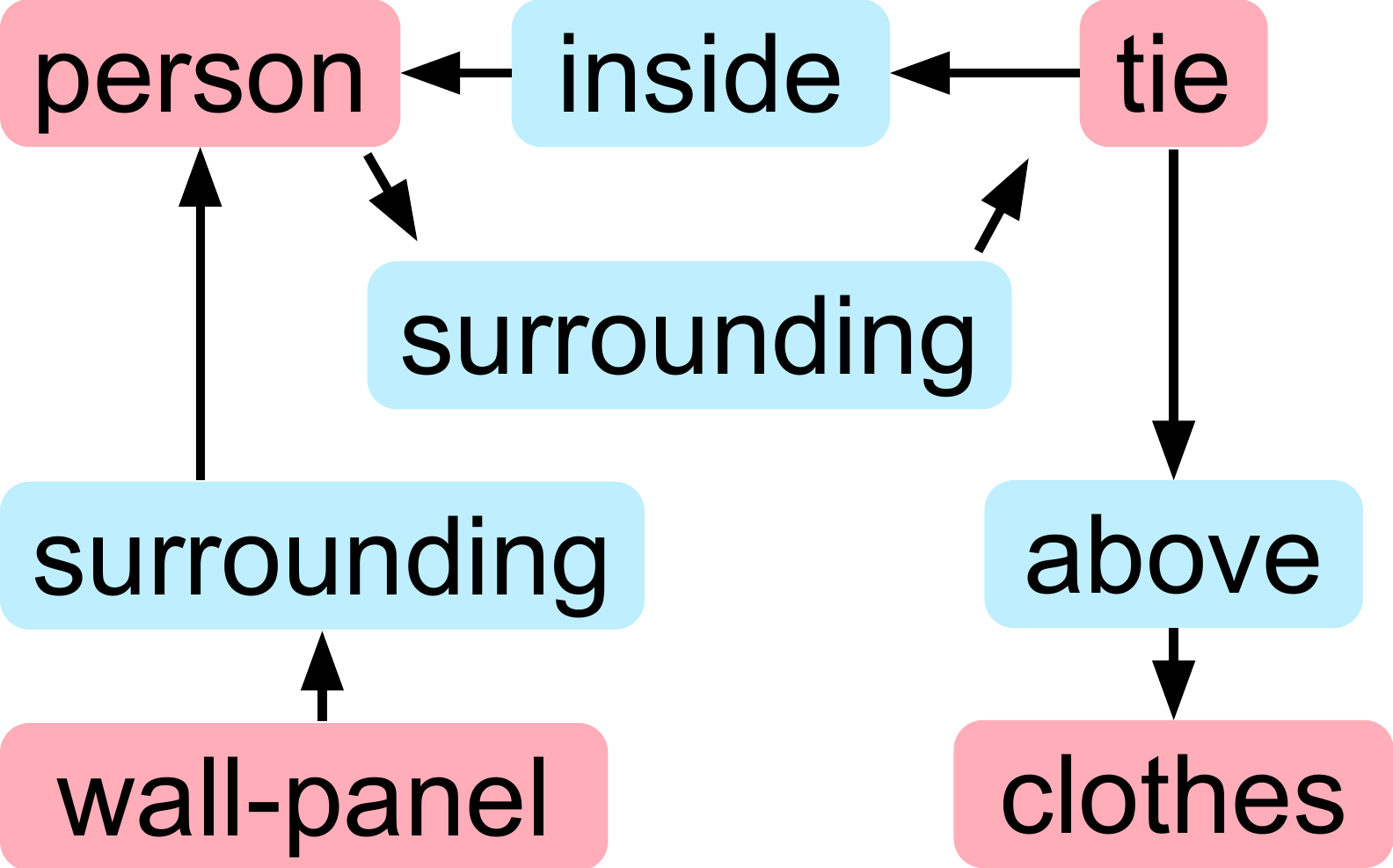}
  \includegraphics[width=\qualsize]{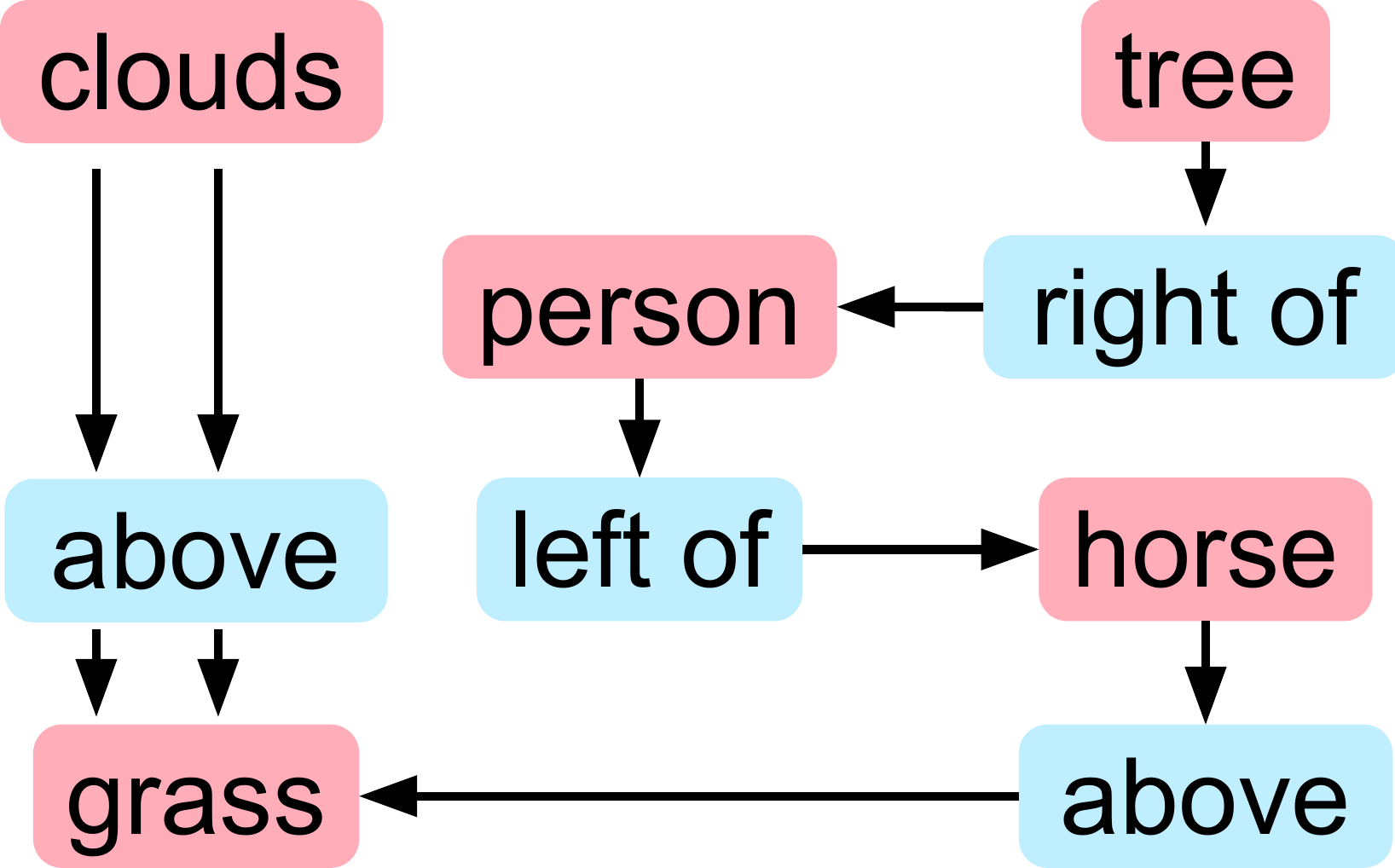}
  \includegraphics[width=\qualsize]{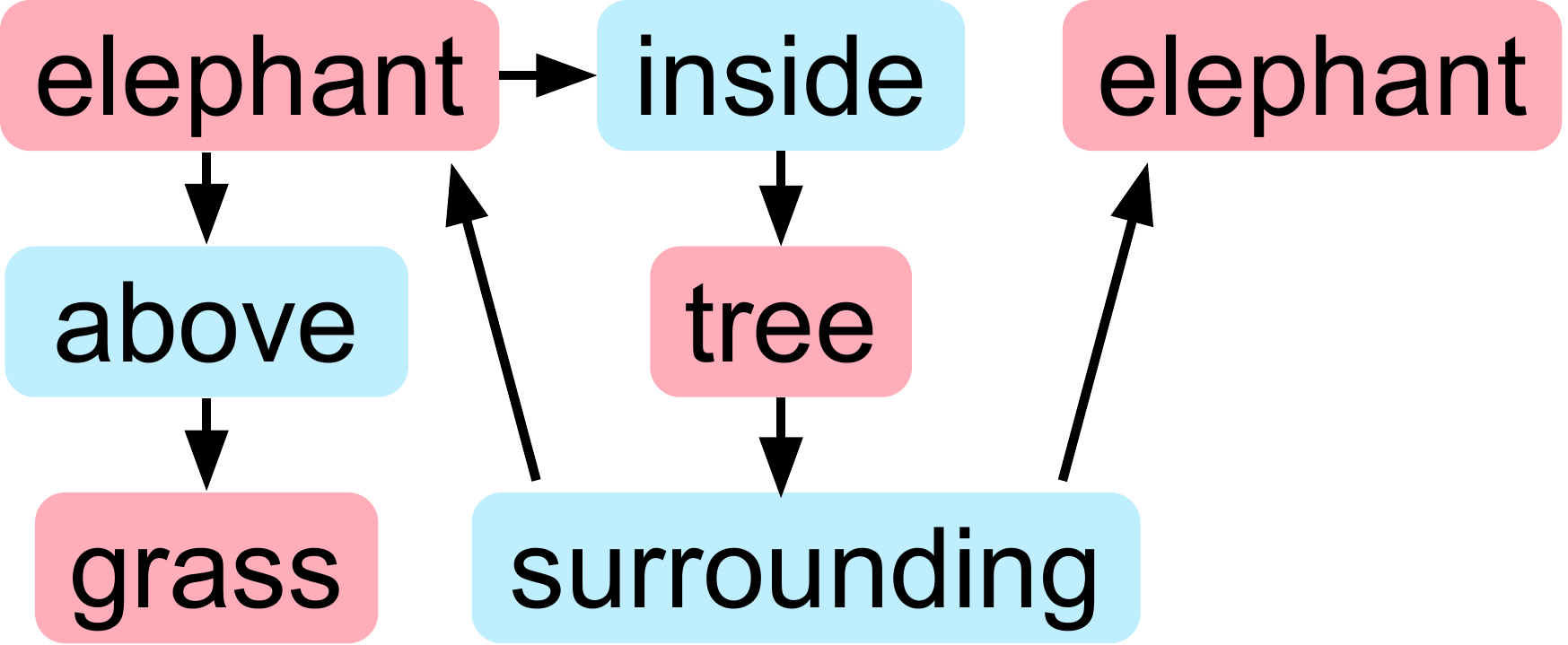}
  \includegraphics[width=\qualsize]{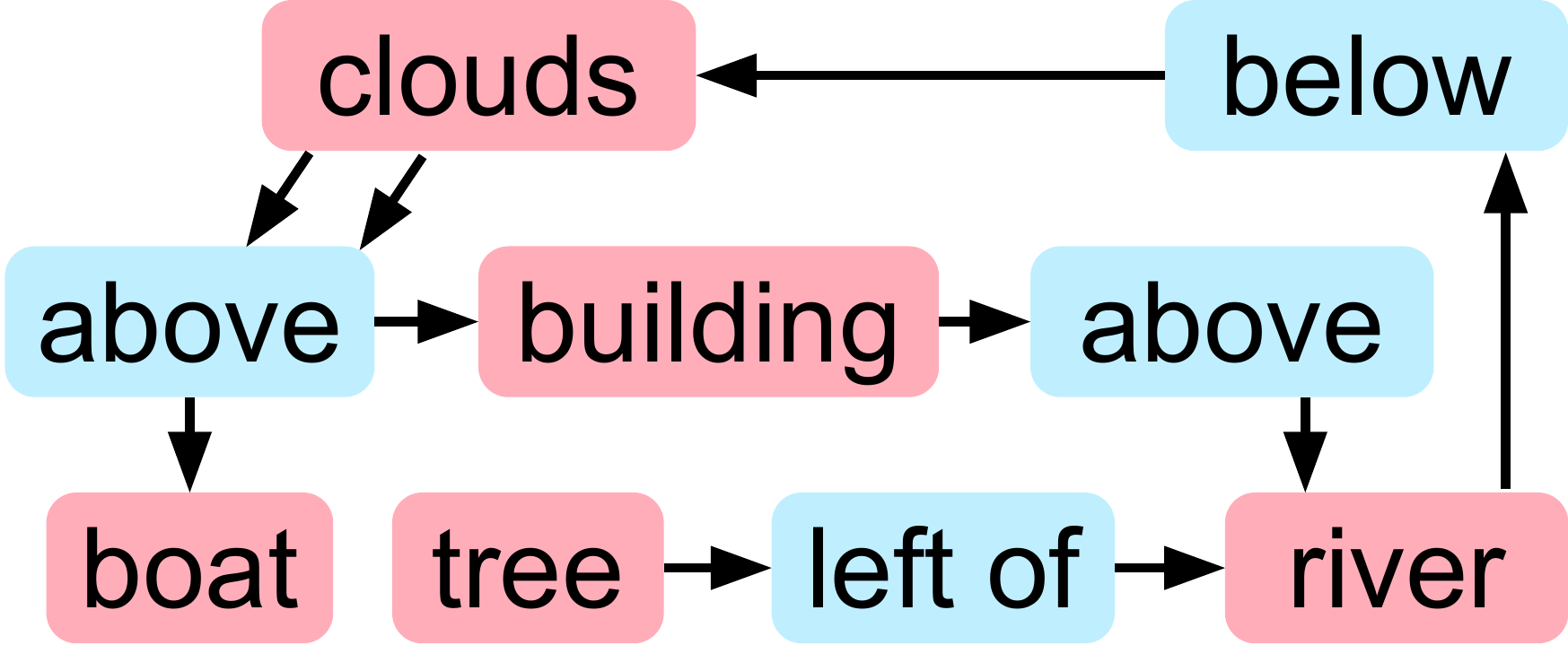} \\*[0.5mm]
  \begin{rotate}{90}
    \hspace{-2mm}\textbf{Text}
  \end{rotate}
  \hspace{\qualtextspace}
  \begin{minipage}{\qualtext}
    \ssmall
    Two cars, one parked on a street with a tree along it, and
    a window in front of a house and a house with a roof.
  \end{minipage}
  \hspace{\qualtextspace}
  \begin{minipage}{\qualtext}
    \ssmall
    Sky above a man riding a horse; the man has a leg and the horse
    has a leg and a tail.
  \end{minipage}
  \hspace{\qualtextspace}
  \begin{minipage}{\qualtext}
    \ssmall
    A boat on top of water; there is also sky, rock, and a bird.
  \end{minipage}
  \hspace{\qualtextspace}
  \begin{minipage}{\qualtext}
    \ssmall
    A glass by a plate with food on it, and another glass by a plate.
  \end{minipage}
  \hspace{\qualtextspace}
  \begin{minipage}{\qualtext}
    \ssmall
    A tie above clothes and inside a person, with a wall panel surrounding
    the person.
  \end{minipage}
  \hspace{\qualtextspace}
  \begin{minipage}{\qualtext}
    \ssmall
    A tree right of a person left of a horse above grass, with clouds
    above the grass.
  \end{minipage}
  \hspace{\qualtextspace}
  \begin{minipage}{\qualtext}
    \ssmall
    An elephant above grass and inside trees surrounding another elephant.
  \end{minipage}
  \hspace{\qualtextspace}
  \begin{minipage}{\qualtext}
    \ssmall
    Clouds above a boat and a building above a river, with trees left of
    the river.
  \end{minipage} \\
  \begin{rotate}{90}
    \hspace{4mm}
    \textbf{Layout}
  \end{rotate}
  \hspace*{0.5mm}
  \includegraphics[width=\qualsize]{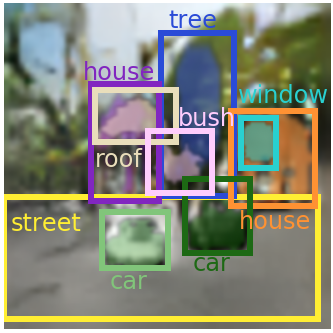}
  \includegraphics[width=\qualsize]{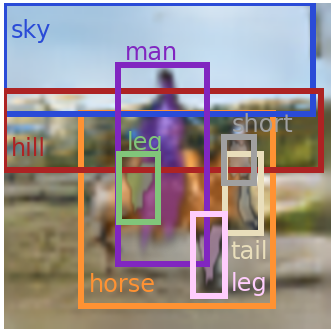}
  \includegraphics[width=\qualsize]{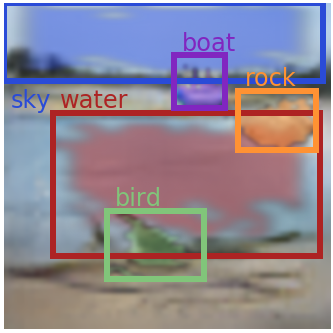}
  \includegraphics[width=\qualsize]{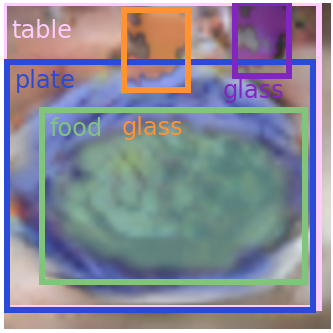}
  \includegraphics[width=\qualsize]{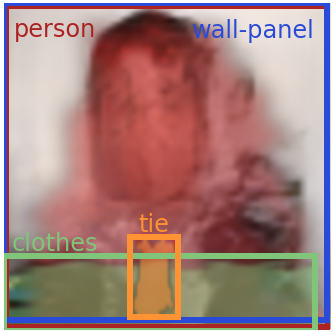}
  \includegraphics[width=\qualsize]{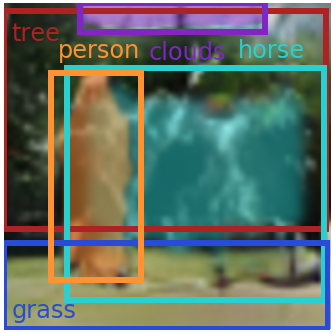}
  \includegraphics[width=\qualsize]{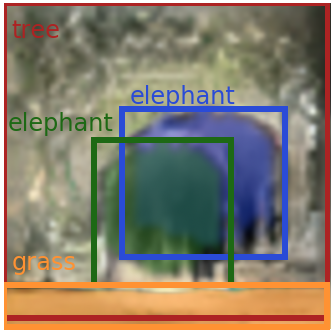}
  \includegraphics[width=\qualsize]{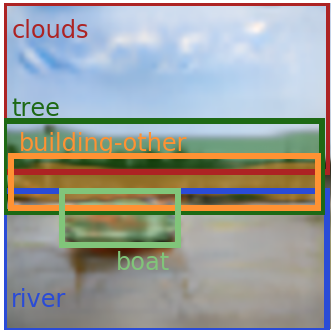} \\
  \begin{rotate}{90}
    \hspace{6mm}\textbf{Image}
  \end{rotate}
  \hspace*{0.5mm}
  \includegraphics[width=\qualsize]{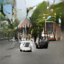}
  \includegraphics[width=\qualsize]{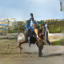}
  \includegraphics[width=\qualsize]{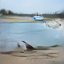}
  \includegraphics[width=\qualsize]{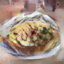} 
  \includegraphics[width=\qualsize]{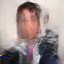}
  \includegraphics[width=\qualsize]{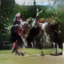}
  \includegraphics[width=\qualsize]{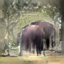}
  \includegraphics[width=\qualsize]{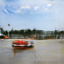} \\
  \begin{rotate}{90}
    \hspace{1mm}
    \textbf{GT Layout}
  \end{rotate}
  \hspace*{0.5mm}
  \includegraphics[width=\qualsize]{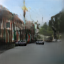}
  \includegraphics[width=\qualsize]{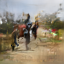}
  \includegraphics[width=\qualsize]{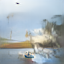}
  \includegraphics[width=\qualsize]{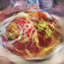} 
  \includegraphics[width=\qualsize]{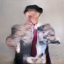}
  \includegraphics[width=\qualsize]{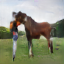}
  \includegraphics[width=\qualsize]{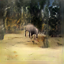}
  \includegraphics[width=\qualsize]{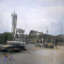} \\*[-1.5mm]
  \begin{minipage}{\qualsize} \centering\scriptsize (i) \end{minipage}
  \begin{minipage}{\qualsize} \centering\scriptsize (j) \end{minipage}
  \begin{minipage}{\qualsize} \centering\scriptsize (k) \end{minipage}
  \begin{minipage}{\qualsize} \centering\scriptsize (l) \end{minipage}
  \begin{minipage}{\qualsize} \centering\scriptsize (m) \end{minipage}
  \begin{minipage}{\qualsize} \centering\scriptsize (n) \end{minipage}
  \begin{minipage}{\qualsize} \centering\scriptsize (o) \end{minipage}
  \begin{minipage}{\qualsize} \centering\scriptsize (p) \end{minipage} \\

  \caption{
    Examples of $64\times64$ generated images using graphs from the test sets of
    Visual Genome (left four columns) and COCO (right four columns).
    For each example we show the input scene graph and a manual translation of the
    scene graph into text; our model processes the scene graph and predicts a layout
    consisting of bounding boxes and segmentation masks for all objects; this layout
    is then used to generate the image.
    We also show some results for our model using ground-truth rather than predicted scene layouts.
    Some scene graphs have duplicate relationships, shown as double arrows.
    For clarity, we omit masks for some stuff categories such as sky, street,
    and water.
  }
  \vspace{-4mm}
  \label{fig:qual}
\end{figure*}

The \emph{object discriminator} $D_{obj}$ ensures that each object in the image
appears realistic; its input are the pixels of an object, cropped and rescaled to
a fixed size using bilinear interpolation~\cite{jaderberg2015spatial}. In addition
to classifying each object as real or fake, $D_{obj}$ also ensures that each object
is recognizable using an \emph{auxiliary classifier}~\cite{odena2017conditional}
which predicts the object's category; both $D_{obj}$ and $f$ attempt to maximize
the probability that $D_{obj}$ correctly classifies objects.

\newlength{\progsize}
\setlength{\progsize}{0.1335\textwidth}
\begin{figure*}[ht]
  \centering
  \begin{minipage}{\progsize}
    \centering
    \scriptsize
    car on street \\*
    line on street \\*
    sky above street
  \end{minipage}
  \begin{minipage}{\progsize}
    \centering
    \scriptsize
    \underline{bus} on street \\*
    line on street \\*
    sky above street
  \end{minipage}
  \begin{minipage}{\progsize}
    \centering
    \scriptsize
    \underline{car on street} \\*
    bus on street \\*
    line on street \\*
    sky above street
  \end{minipage}
  \begin{minipage}{\progsize}
    \centering
    \scriptsize
    car on street \\*
    bus on street \\*
    line on street \\*
    sky above street \\*
    \underline{kite in sky}
  \end{minipage}
  \begin{minipage}{\progsize}
    \centering
    \ssmall
    car on street \\*
    bus on street \\*
    line on street \\*
    sky above street \\*
    kite in sky \\*
    \underline{car below kite}
  \end{minipage}
  \begin{minipage}{\progsize}
    \centering
    \ssmall
    car on street \\*
    bus on street \\*
    line on street \\*
    sky above street \\*
    \underline{building behind street}
  \end{minipage}
  \begin{minipage}{\progsize}
    \centering
    \ssmall
    car on street \\*
    bus on street \\*
    line on street \\*
    sky above street \\*
    building behind street \\*
    \underline{window on building}
  \end{minipage} \\
  \includegraphics[width=\progsize]{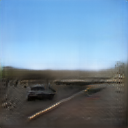}
  \includegraphics[width=\progsize]{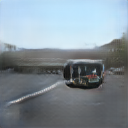}
  \includegraphics[width=\progsize]{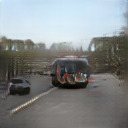}
  \includegraphics[width=\progsize]{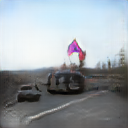}
  \includegraphics[width=\progsize]{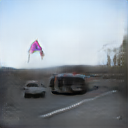}
  \includegraphics[width=\progsize]{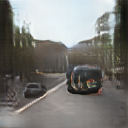}
  \includegraphics[width=\progsize]{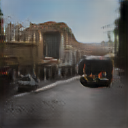} \\*[1mm]
  \begin{minipage}{\progsize}
    \centering
    \scriptsize
    sky above grass \\*
    zebra standing on grass
  \end{minipage}
  \begin{minipage}{\progsize}
    \centering
    \scriptsize
    sky above grass \\*
    \underline{sheep} standing on grass
  \end{minipage}
  \begin{minipage}{\progsize}
    \centering
    \scriptsize
    sky above grass \\*
    sheep standing on grass \\*
    \underline{sheep' by sheep}
  \end{minipage}
  \begin{minipage}{\progsize}
    \centering
    \ssmall
    sky above grass \\*
    sheep standing on grass \\*
    sheep' by sheep \\*
    \underline{tree behind sheep}
  \end{minipage}
  \begin{minipage}{\progsize}
    \centering
    \ssmall
    sky above grass \\*
    sheep standing on grass \\*
    tree behind sheep \\*
    sheep' by sheep \\*
    \underline{ocean by tree}
  \end{minipage}
  \begin{minipage}{\progsize}
    \centering
    \ssmall
    sky above grass \\*
    sheep standing on grass \\*
    tree behind sheep \\*
    sheep' by sheep \\*
    ocean by tree \\*
    \underline{boat in ocean}
  \end{minipage}
  \begin{minipage}{\progsize}
    \centering
    \ssmall
    sky above grass \\*
    sheep standing on grass \\*
    tree behind sheep \\*
    sheep' by sheep \\*
    ocean by tree \\*
    boat \underline{on grass}
  \end{minipage} \\*
  \includegraphics[width=\progsize]{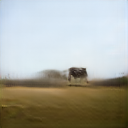}
  \includegraphics[width=\progsize]{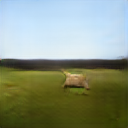}
  \includegraphics[width=\progsize]{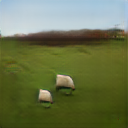}
  \includegraphics[width=\progsize]{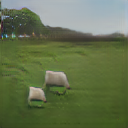}
  \includegraphics[width=\progsize]{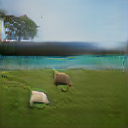}
  \includegraphics[width=\progsize]{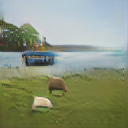}
  \includegraphics[width=\progsize]{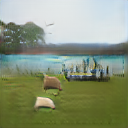}
  \caption{
    Images generated by our method trained on Visual Genome. In each row we
    start from a simple scene graph on the left and progressively add more
    objects and relationships moving to the right. Images respect
    relationships like \emph{car below kite} and \emph{boat on grass}.
  }
  \vspace{-4mm}
  \label{fig:manual}
\end{figure*}

\textbf{Training.}
We jointly train the generation network $f$ and the discriminators $D_{obj}$
and $D_{img}$. The generation network is trained to minimize the weighted sum of
six losses:

\begin{itemize}[leftmargin=*]
  \setlength\itemsep{-1mm}
  \item \emph{Box loss} $\LL_{box}=\sum_{i=1}^n\|b_i - \hat b_i\|_1$
    penalizing the $L_1$ difference between ground-truth and predicted boxes
  \item \emph{Mask loss} $\LL_{mask}$ penalizing differences between
    ground-truth and predicted masks with pixelwise
    cross-entropy; not used for models trained on Visual Genome
  \item \emph{Pixel loss} $\LL_{pix}=\|I - \hat I\|_1$ penalizing the
    $L_1$ difference between ground-truth generated images
  \item \emph{Image adversarial loss} $\LL_{GAN}^{img}$ from $D_{img}$
    encouraging generated image patches to appear realistic
  \item \emph{Object adversarial loss} $\LL_{GAN}^{obj}$ from the $D_{obj}$
    encouraging each generated object to look realistic
  \item \emph{Auxiliarly classifier loss} $\LL_{AC}^{obj}$ from $D_{obj}$,
    ensuring that each generated object can be classified by $D_{obj}$
\end{itemize}

\textbf{Implementation Details.}
We augment all scene graphs with a special \emph{image} object, and add
special \emph{in image} relationships connecting each true object with the
\emph{image} object; this ensures that all scene graphs are connected.

We train all models using Adam~\cite{kingma2015adam} with learning rate
$10^{-4}$ and batch size 32 for 1 million iterations; training
takes about 3 days on a single Tesla P100.
For each minibatch we first update $f$, then update $D_{img}$ and $D_{obj}$.

We use ReLU for graph convolution; the CRN and discriminators use 
discriminators use LeakyReLU~\cite{maas2013rectifier} and batch
normalization~\cite{ioffe2015batch}. Full details about our architecture can
be found in the supplementary material, and code will be made publicly available.

\section{Experiments}
We train our model to generate $64\times64$ images on the Visual Genome~\cite{krishna2017visual} and COCO-Stuff~\cite{caesar2016coco} datasets. In our experiments
we aim to show that our method generates images of complex scenes which respect
the objects and relationships of the input scene graph.

\subsection{Datasets}

\textbf{COCO.}
We perform experiments on the 2017 COCO-Stuff dataset~\cite{caesar2016coco},
which augments a subset of the COCO dataset~\cite{lin2014microsoft} with
additional stuff categories.  The dataset annotates 40K train and 5K val images
with bounding boxes and segmentation masks for 80 \emph{thing} categories
(people, cars, \etc) and 91 \emph{stuff} categories (sky, grass, \etc).

We use these annotations to construct synthetic scene graphs based on the 2D
image coordinates of the objects, using six mutually exclusive geometric
relationships: \emph{left of}, \emph{right of}, \emph{above}, \emph{below},
\emph{inside}, and \emph{surrounding}.

We ignore objects covering less than 2\% of the image, and use images with 3 to
8 objects; we divide the COCO-Stuff 2017 val set into our own val and test sets,
leaving us with 24,972 train, 1024 val, and 2048 test images.

\textbf{Visual Genome.}
We experiment on Visual Genome~\cite{krishna2017visual} version 1.4 (VG) which
comprises 108,077 images annotated with scene graphs. We divide the data into
80\% train, 10\% val, and 10\% test; we use object and relationship categories
occurring at least 2000 and 500 times respectively in the train set, leaving
178 object and 45 relationship types.

We ignore small objects,
and use images with between 3 and 30 objects and at
least one relationship; this leaves us with 62,565 train, 5,506 val, and 5,088
test images with an average of ten objects and five relationships per image.

Visual Genome does not provide segmentation masks, so we omit the
mask prediction loss for models trained on VG.

\subsection{Qualitative Results}
Figure~\ref{fig:qual} shows example scene graphs from the Visual Genome and COCO
test sets and generated images using our method, as well as predicted object
bounding boxes and segmentation masks.

From these examples it is clear that our method can generate scenes with multiple
objects, and even multiple instances of the same object type: for example Figure~\ref{fig:qual} (a) shows two sheep, (d) shows two busses, (g) contains three 
people, and (i) shows two cars.

These examples also show that our method generates images which respect the
relationships of the input graph; for example in (i) we see one broccoli 
\emph{left of} a second broccoli, with a carrot \emph{below} the second broccoli; 
in (j) the man is \emph{riding} the horse, and both the man and the horse have
\emph{legs} which have been properly positioned.

Figure~\ref{fig:qual} also shows examples of images generated by our method
using ground-truth rather than predicted object layouts. In some cases we see
that our predicted layouts can vary significantly from the ground-truth objects
layout. For example in (k) the graph does not specify the position of the bird
and our method renders it standing on the ground, but in the ground-truth layout 
the bird is flying in the sky. Our model is sometimes bottlenecked by layout
prediction, such as (n) where using the ground-truth rather than predicted
layout significantly improves the image quality.

In Figure~\ref{fig:manual} we demonstrate our model's ability to generate complex
images by starting with simple graphs on the left and progressively building up
to more complex graphs. From this example we can see that object positions are
influenced by the relationships in the graph: in the top sequence
adding the relationship \emph{car below kite} causes the car to shift to the
right and the kite to shift to the left so that the relationship is respected.
In the bottom sequence, adding the relationship \emph{boat on grass} causes the
boat's position to shift.

\begin{table}
  \setlength{\tabcolsep}{2mm}
  \centering
  \scalebox{0.9}{
  \begin{tabular}{l|cc}
                               & \multicolumn{2}{c}{\textbf{Inception}}   \\
    \textbf{Method}            & COCO & VG \\
    \hline\hline               
    Real Images $(64\times64)$ & $16.3 \pm 0.4$ & $13.9 \pm 0.5$ \\
    \hline                     
    Ours (No gconv)            & $4.6 \pm 0.1$ & $4.2 \pm 0.1$ \\
    Ours (No relationships)    & $3.7 \pm 0.1$ & $4.9 \pm 0.1$ \\
    Ours (No discriminators)   & $4.8 \pm 0.1$ & $3.6 \pm 0.1$ \\
    Ours (No $D_{obj}$)        & $5.6 \pm 0.1$ & $5.0 \pm 0.2$ \\
    Ours (No $D_{img}$)        & $5.6 \pm 0.1$ & $\mathbf{5.7 \pm 0.3}$ \\
    Ours (Full model)          & $\mathbf{6.7 \pm 0.1}$ & $5.5 \pm 0.1$ \\
    \hline                     
    Ours (GT Layout, no gconv) & $7.0 \pm 0.2$ & $6.0 \pm 0.2$ \\
    Ours (GT Layout)           & $\mathbf{7.3 \pm 0.1}$ & $\mathbf{6.3 \pm 0.2}$ \\
    \hline
    StackGAN~\cite{zhang2017stackgan} $(64\times64)$ & $\mathbf{8.4 \pm 0.2}$ & - \\
  \end{tabular}}
  \vspace{1mm}
  \caption{
    Ablation study using Inception scores. On each dataset we randomly 
    split our test-set samples into 5 groups and report mean and standard
    deviation across splits. On COCO we generate five samples for each test-set
    image by constructing different synthetic scene graphs.
    For StackGAN we generate one image for each of the COCO test-set captions, and
    downsample their $256\times256$ output to $64\times64$ for fair comparison with
    our method.
  }
  \vspace{-2mm}
  \label{tab:inception}
\end{table}

\subsection{Ablation Study}
We demonstrate the necessity of all components of our model by comparing the
image quality of several ablated versions of our model, shown in Table~\ref{tab:inception}; see supplementary material for example images from ablated models.

We measure image quality using \emph{Inception score}\footnote{Defined as $\exp(\mathbb{E}_{\hat I} KL(p(y|\hat I)\|p(y)))$ where the expectation is taken over generated images $\hat I$ and $p(y|\hat I)$ is the predicted label distribution.}~\cite{salimans2016improved}
which uses an ImageNet classification model~\cite{russakovsky2015imagenet,szegedy2015going}
to encourage recognizable objects within images and diversity across images.
We test several ablations of our model:

\textbf{No gconv} omits graph convolution, so boxes and
masks are predicted from initial object embedding vectors.
It cannot reason jointly about the presence of different objects, and can only predict one box and mask per category.

\textbf{No relationships} uses graph convolution layers but ignores all
relationships from the input scene graph except for trivial \emph{in image}
relationships; graph convolution allows this model to jointly about
objects. Its poor performance demonstrates the utility of the scene graph relationships.

\textbf{No discriminators} omits both $D_{img}$ and $D_{obj}$, relying on the
pixel regression loss $\LL_{pix}$ to guide the generation network.
It tends to produce overly smoothed images.

\textbf{No} $\mathbf{D_{obj}}$ and \textbf{No} $\mathbf{D_{img}}$ omit one of
the discriminators. On both datasets, using any discriminator leads
to significant improvements over models trained with $\LL_{pix}$ alone.
On COCO the two discriminators are complimentary, and combining
them in our full model leads to large improvements.
On VG, omitting $D_{img}$ does not degrade performance.

In addition to ablations, we also compare with two \textbf{GT Layout}
versions of our model which omit the $\LL_{box}$ and $\LL_{mask}$ losses, and
use ground-truth bounding boxes during both training and testing; on COCO they
also use ground-truth segmentation masks, similar to Chen and
Koltun~\cite{chen2017photographic}. These methods give an upper bound to our
model's performance in the case of perfect layout prediction.

Omitting graph convolution degrades performance even when using ground-truth layouts, suggesting that scene graph relationships and graph convolution have benefits beyond simply predicting object positions.

\begin{table}
  \centering
  \setlength{\tabcolsep}{0.75mm}
  \scalebox{0.82}{
    \begin{tabular}{c|cc|cc|cc|cc}
      & \multicolumn{2}{c|}{R@0.3} & \multicolumn{2}{c|}{R@0.5} & \multicolumn{2}{c|}{$\sigma_x$} & \multicolumn{2}{c}{$\sigma_{area}$} \\
      & COCO & VG & COCO & VG & COCO & VG & COCO & VG \\
      \hline\hline
      Ours (No gconv) & 46.9 & 20.2 & 20.8 &  6.4 & 0   & 0 & 0 & 0 \\
      Ours (No rel.)  & 21.8 & 16.5 &  7.6 &  6.9 & \textbf{0.1} & \textbf{0.1} & \textbf{0.2} & \textbf{0.1} \\
      Ours (Full)     & \textbf{52.4} &\textbf{21.9} & \textbf{32.2} & \textbf{10.6} & \textbf{0.1} & \textbf{0.1} & \textbf{0.2} & \textbf{0.1} \\
    \end{tabular}
  }
  \vspace{1mm}
  \caption{
    Statistics of predicted bounding boxes.
    R@$t$ is object recall with an IoU threshold of $t$, and measures agreement with ground-truth boxes.
    $\sigma_x$ and $\sigma_{area}$ measure box variety by computing the standard deviation of box $x$-positions and areas within
    each object category and then averaging across categories.
  }
  \vspace{-2mm}
  \label{tab:bbox}
\end{table}

\subsection{Object Localization}
In addition to looking at images, we can also inspect
the bounding boxes predicted by our model. One measure of box quality is
high agreement between predicted and ground-truth boxes; in
Table~\ref{tab:bbox} we show the object recall of our model at two
intersection-over-union thresholds.

Another measure for boxes is \emph{variety}: predicted boxes for objects 
should vary in response to the other objects and relationships in the graph.
Table~\ref{tab:bbox} shows the mean
per-category standard deviations of box position and area.

Without graph convolution, our model can only learn to predict a single bounding
box per object category. This model achieves nontrivial object recall, but has no
variety in its predicted boxes, as $\sigma_x=\sigma_{area}=0$.

Using graph convolution without relationships, our model can jointly reason about
objects when predicting bounding boxes; this leads to improved variety in its
predictions. Without relationships, this model's predicted boxes have
less agreement with ground-truth box positions.

Our full model with graph convolution and relationships achieves both variety and 
high agreement with ground-truth boxes, indicating that it can use the
relationships of the graph to help localize objects with greater fidelity.

\begin{figure}
  \centering
  \begin{minipage}{0.09\textwidth}
    \centering
    \textbf{Caption}
  \end{minipage}
  \begin{minipage}{0.11\textwidth}
    \centering
    \textbf{StackGAN~\cite{zhang2017stackgan}}
  \end{minipage}
  \begin{minipage}{0.11\textwidth}
    \centering
    \textbf{Ours}
  \end{minipage}
  \begin{minipage}{0.115\textwidth}
    \centering
    \textbf{Scene Graph}
  \end{minipage} \\*
  \begin{minipage}[b]{0.09\textwidth}
    \centering
    \footnotesize
    \textit{A person skiing down a slope next to snow covered trees}
    \vspace{2mm}
  \end{minipage}
  \begin{minipage}[t]{0.11\textwidth}
    \includegraphics[width=\textwidth]{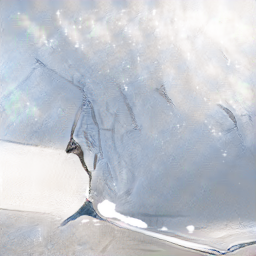}
  \end{minipage}
  \begin{minipage}[t]{0.11\textwidth}
    \includegraphics[width=\textwidth]{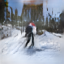}
  \end{minipage}
  \begin{minipage}[b]{0.115\textwidth}
    \centering
    \includegraphics[height=20mm]{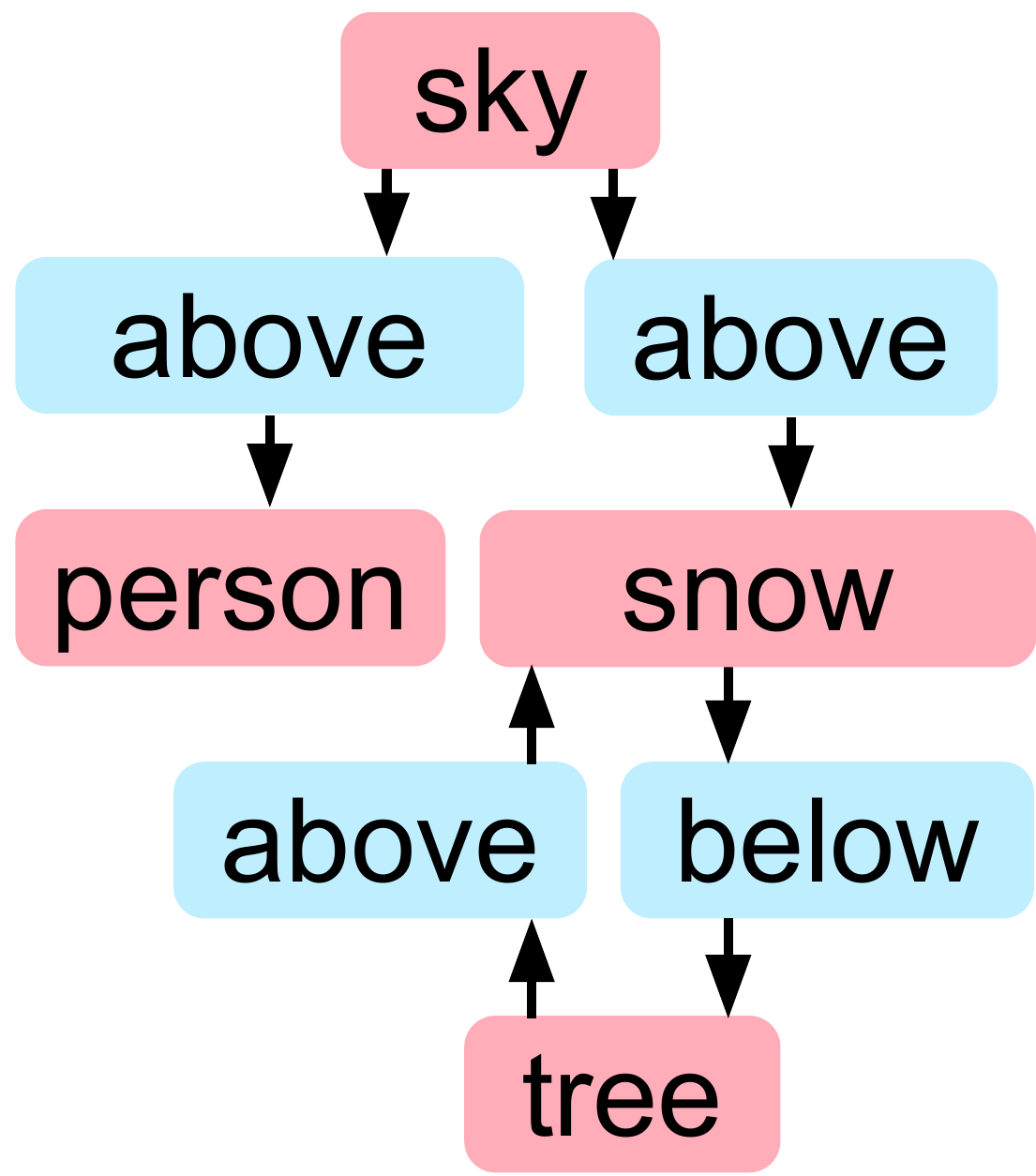}
  \end{minipage} \\*[0.5mm]
  {\footnotesize
    \hspace{-4mm} \emph{Which image matches the caption better?} \\*[1mm]
  }
  \hspace*{-14.5mm}\begin{tabular}{c|c|c}
    \hline
    \textbf{User}  & \hspace{1mm} 332 / 1024 \hspace{1mm} & \hspace{1mm} \textbf{692} / 1024 \hspace{1mm} \\
    \textbf{choice}  & (32.4\%) & \textbf{(67.6\%)}
  \end{tabular}
  \vspace{1mm}
  \caption{
    We performed a user study to compare the semantic interpretability of our method against
    StackGAN~\cite{zhang2017stackgan}. \textbf{Top:} We use StackGAN to generate an image from a
    COCO caption, and use our method to generate an image from a scene graph constructed from the
    COCO objects corresponding to the caption. We show users the caption and both images,
    and ask which better matches the caption. \textbf{Bottom:} Across 1024 val image pairs, users prefer the
    results from our method by a large margin.
  }
  \vspace{-4mm}
  \label{fig:caption-study}
\end{figure}

\subsection{User Studies}
\label{sec:user-study}
Automatic metrics such as Inception scores and box statistics give a
coarse measure of image quality; the true measure of success is
human judgement of the generated images. For this reason we performed two user
studies on Mechanical Turk to evaluate our results.

We are unaware of any previous end-to-end methods for generating images from
scene graphs, so we compare our method with StackGAN~\cite{zhang2017stackgan},
a state-of-the art method for generating images from sentence descriptions.

Despite the different input modalities between our method and StackGAN, we can
compare the two on COCO, which in addition to object annotations also provides
captions for each image.
We use our method to generate images from synthetic scene
graphs built from COCO object annotations, and StackGAN\footnote{We use the pretrained COCO model provided by the authors at \url{https://github.com/hanzhanggit/StackGAN-Pytorch}} to generate images from COCO captions for the same images.
Though the methods receive different inputs, they should generate similar images due to the correspondence between COCO captions and objects.

For user studies we downsample StackGAN images to $64\times64$ to compensate for differing resolutions; we repeat all trials with three workers and randomize order in all trials.

\textbf{Caption Matching.}
We measure semantic interpretability by showing users a COCO caption, an image generated by StackGAN from that caption, and an image generated by our method from a scene graph built from the COCO objects corresponding to the caption. We ask users to select the image that better matches the caption. An example image pair and results are shown in Figure~\ref{fig:caption-study}.

This experiment is biased toward StackGAN, since the caption may contain
information not captured by the scene graph. Even so, a majority of workers
preferred the result from our method in 67.6\% of image pairs,
demonstrating that compared to StackGAN our method more
frequently generates complex, semantically meaningful images.

\begin{figure}
  \centering
  \begin{minipage}{0.09\textwidth}
    \centering
    \textbf{Caption}
  \end{minipage}
  \begin{minipage}{0.11\textwidth}
    \centering
    \textbf{StackGAN~\cite{zhang2017stackgan}}
  \end{minipage}
  \begin{minipage}{0.11\textwidth}
    \centering
    \textbf{Ours}
  \end{minipage}
  \begin{minipage}{0.115\textwidth}
    \centering
    \textbf{Scene Graph}
  \end{minipage} \\*
  \begin{minipage}[b]{0.09\textwidth}
    \centering
    \footnotesize
    \textit{A \underline{man} flying through the \underline{air} while riding a \underline{bike}.}
    \vspace{2mm}
  \end{minipage}
  \begin{minipage}[b]{0.11\textwidth}
    \includegraphics[width=\textwidth]{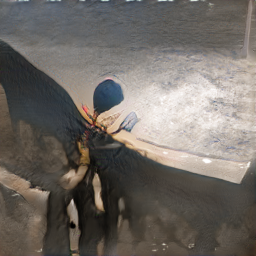}
  \end{minipage}
  \begin{minipage}[b]{0.11\textwidth}
    \includegraphics[width=\textwidth]{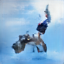}
  \end{minipage}
  \begin{minipage}[b]{0.115\textwidth}
    \centering
    \includegraphics[height=19mm]{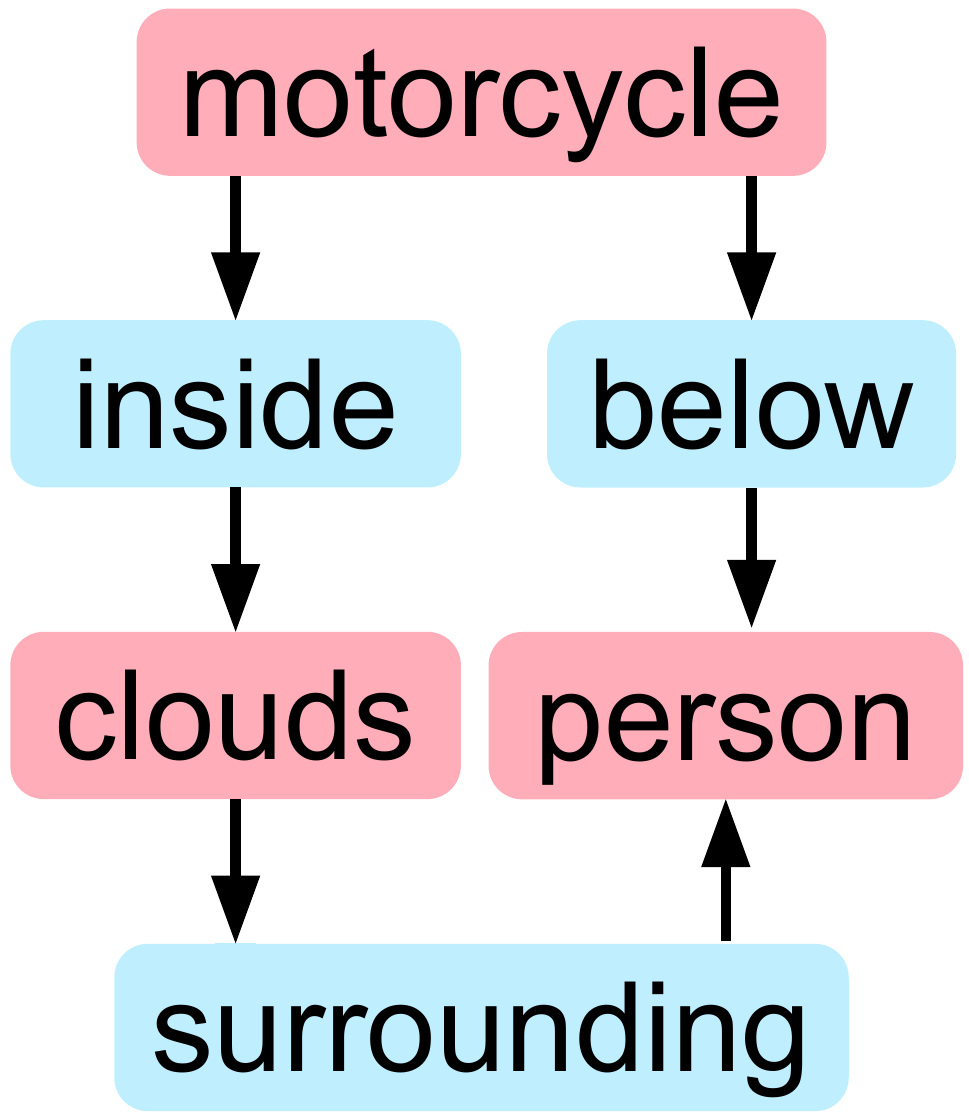}
  \end{minipage} \\*[0.5mm]
  {\footnotesize
    \hspace{-16mm}\textit{Which objects are present?} motorcycle, person, clouds \\*[1mm]}
  \hspace*{-17mm}\begin{tabular}{c|cc}
    \hline
    \textbf{Thing}   & \hspace{1mm} 470 / 1650 \hspace{1mm} & \hspace{1mm} \textbf{772} / 1650 \hspace{1mm} \\
    \textbf{recall}  & (28.5\%)   & \textbf{(46.8\%)} \\
    \hline
    \textbf{Stuff}   & 1285 / 3556 & \textbf{2071} / 3556 \\
    \textbf{Recall}  & (36.1\%)    & \textbf{(58.2\%)}
  \end{tabular}
  \vspace{1mm}
  \caption{
    We performed a user study to measure the number of recognizable objects in images from our
    method and from StackGAN~\cite{zhang2017stackgan}. \textbf{Top:} We use StackGAN to generate
    an image from a COCO caption, and use our method to generate an image from a scene graph built from
    the COCO objects corresponding to the caption. For each image, we ask users which COCO objects they
    can see in the image. \textbf{Bottom:} Across 1024 val image pairs, we measure the fraction of \emph{things}
    and \emph{stuff} that users can recognize in images from each method. Our method produces more objects.
  }
    \label{fig:object-study}
    \vspace{-3mm}
\end{figure}

\textbf{Object Recall.}
This experiment measures the number of recognizable objects in each method's images. In each trial we show an image from one method and a list of COCO objects and ask users to identify which objects appear in the image. An example and results are snown in Figure~\ref{fig:object-study}.

We compute the fraction of objects that a majority of users believed were
present, dividing the results into \emph{things} and \emph{stuff}. Both methods
achieve higher recall for stuff than things, and our method achieves
significantly higher object recall, with 65\% and 61\% relative improvements for thing and stuff recall respectively.

This experiment is biased toward our method since the scene graph may
contain objects not mentioned in the caption, but it demonstrates
that compared to StackGAN, our method produces images with
more recognizable objects.

\section{Conclusion}
In this paper we have developed an end-to-end method for generating images from
scene graphs. Compared to leading methods which generate images from text
descriptions, generating images from structured scene graphs rather than
unstructured text allows our method to reason explicitly about objects and
relationships, and generate complex images with many recognizable objects.

\textbf{Acknowledgments} We thank Shyamal Buch, Christopher Choy, De-An Huang, and Ranjay Krishna for helpful comments and suggestions.

{\small
\bibliographystyle{ieee}
\bibliography{refs}
}

\clearpage
\appendix
\begin{figure}[ht!]
  \centering
  \Large\textbf{Supplementary Material}
\end{figure}
\section{Network Architecture}
Here we describe the exact network architectures for all components of our model.

\subsection{Graph Convolution Layer}
As described in Section 3 of the main paper, we process the input scene graph with a \emph{graph convolution network} composed of several \emph{graph convolution layers}.

A graph convolution layer accepts as input a vector of dimension $D_{in}$ for each node and edge in the graph, and computes new vectors of dimension $D_{out}$ for each node and edge. A single graph convolution layer can be applied to graphs of any size of shape due to \emph{weight sharing}. A single graph convolution layer proceeds in two stages.

First, along relationship of the scene graph we apply three functions $g_s$, $g_p$, and $g_o$; these functions take as input the vectors $v_s$, $v_r$, and $v_o$ for the starting node, edge, and ending node of the relationship and produce new vectors for the two nodes and the edge. The new vector for the edge $v_r'=g_p(v_s, v_r, v_o)$ has dimension $D_{out}$, and is used as the output vector from the graph convolution layer for the edge. The new vectors for the starting and ending nodes $\tilde v_s = g_s(v_s, v_r, v_o)$ and $\tilde v_o = g_o(v_s, v_r, v_o)$ are \emph{candidate vectors} of dimension $H$. In practice the three functions $g_s$, $g_p$, and $g_o$ and implemented with a single multilayer perceptron (MLP) whose architecture is shown in Table~\ref{tab:gconv-g}.

\begin{table}
  \centering
  \setlength{\tabcolsep}{1mm}
  \scalebox{0.88}{
  \begin{tabular}{|c|c|c|c|}
    \hline
    \textbf{Index} & \textbf{Inputs} & \textbf{Operation} & \textbf{Output shape} \\
    \hline
    (1) & - & Subject vector $v_s$ & $D_{in}$ \\
    (2) & - & Relationship vector $v_r$ & $D_{in}$ \\
    (3) & - & Object vector $v_o$ & $D_{in}$ \\
    (4) & (1), (2), (3)& Concatenate & $3D_{in}$ \\
    (5) & (4) & Linear($3D_{in}\rightarrow H)$ & $H$ \\
    (6) & (5) & ReLU & $H$ \\
    (7) & (6) & Linear($H \rightarrow 2H + D_{out}$) & $2H + D_{out}$ \\
    (8) & (7) & ReLU & $2H + D_{out}$ \\
    (9) & (8) & Split into 3 chunks & $H, D_{out}, H$ \\
    (10) & (9) & 1st chunk $\tilde v_s$ & $H$ \\
    (11) & (9) & 2nd chunk $v'_r$ & $D_{out}$ \\
    (12) & (9) & 3rd chunk $\tilde v_o$ & $H$ \\
    \hline
  \end{tabular}}
  \vspace{1mm}
  \caption{
    Network architecture for the first network $g$ used in graph convolution;
    this single network implements the three functions $g_s$, $g_p$, and $g_o$
    from the main text.
  }
  \label{tab:gconv-g}
\end{table}

As a second stage of processing, for each object in the scene graph we collect all of its candidate vectors and process them with a symmeitric pooling function $h$ which converts the set of candidate vectors into a a single vector of dimension $D_{out}$. Concretely, for object $o_i$ in the scene graph $G$, let $V_i^s=\{g_s(v_i,v_r,v_j):(o_i,r,o_j)\in G\}$ be the set of candidate vectors for $o_i$ from relationships where $o_i$ appears as the subject, and let $V_i^o=\{g_o(v_j, v_r, v_i):(o_j, r, o_i)\in G\}$ be the set of candidate vectors for $o_i$ from relationships where $o_i$ appears as the object of the relationship. The pooling function $h$ takes as input the two sets of vectors $V_i^s$ and $V_i^o$, averages them, and feeds the result to an MLP to compute the output vector $v_i'$ for object $o_i$ from the graph convolution layer. The exact architecture of the network we use for $h$ is shown in Table~\ref{tab:gconv-h}.

Overall a graph convolution layer has three hyperparameters defining its size: the \emph{input dimension} $D_{in}$, the \emph{hidden dimension} $H$, and the \emph{output dimension} $D_{out}$. We can therefore specify a graph convolution layer with the notation gconv($D_{in}\to H\to D_{out}$).

\begin{table}
  \centering
  \setlength{\tabcolsep}{1mm}
  \scalebox{0.88}{
  \begin{tabular}{|c|c|c|c|}
    \hline
    \textbf{Index} & \textbf{Inputs} & \textbf{Operation} & \textbf{Output Shape} \\
    \hline
    (1) & - & Subject candidate set $V_i^s$ & $|V_i^s| \times H$ \\
    (2) & - & Object candidate set $V_i^o$ & $|V_i^o| \times H$ \\
    (3) & (1), (2) & Set union & $(|V_i^s|+|V_i^o|) \times H$ \\
    (4) & (3) & Mean over axis 0 & $H$ \\
    (5) & (4) & Linear($H\rightarrow H$) & $H$ \\
    (6) & (5) & ReLU & $H$ \\
    (7) & (6) & Linear($H\rightarrow D_{out}$) & $D_{out}$ \\
    (8) & (7) & ReLU & $D_{out}$ \\
    \hline
  \end{tabular}}
  \vspace{1mm}
  \caption{
    Network architecture for the second network $h$ used in graph convolution;
    this network implements a symmetric pooling function to convert the set of
    all candidate vectors for an object into a single output vector.
  }
  \label{tab:gconv-h}
\end{table}

\subsection{Graph Convolution Network}
The input scene graph is processed by a \emph{graph convolution network}, the
exact architecture of which is shown in Table~\ref{tab:gconv-net}. Our network
first embeds the objects and relationships of the graph with embedding layers to
produce vectors of dimension $D_{in}=128$; we then use five layers of graph convolution
with $D_{in}=D_{out}=128$ and $H=512$.

\begin{table}
  \centering
  \setlength{\tabcolsep}{1mm}
  \scalebox{0.85}{
  \begin{tabular}{|c|c|c|c|}
    \hline
    \textbf{Index} & \textbf{Inputs} & \textbf{Operation} & \textbf{Output Shape} \\
    \hline
    (1) & - & Graph objects & $O$ \\
    (2) & - & Graph relationships & $R$ \\
    (3) & (1) & Object Embedding & $O \times 128$ \\
    (4) & (2) & Relationship embedding & $R \times 128$ \\
    (5) & (1), (2) & gconv($128\to512\to128$) & $O\times128, R\times128$ \\
    (6) & (5) & gconv($128\to512\to128$) & $O\times 128$, $R \times 128$ \\
    (7) & (6) & gconv($128\to512\to128$) & $O\times 128$, $R \times 128$ \\
    (8) & (7) & gconv($128\to512\to128$) & $O\times 128$, $R \times 128$ \\
    (9) & (8) & gconv($128\to512\to128$) & $O\times 128$, $R \times 128$ \\
    \hline
  \end{tabular}}
  \vspace{1mm}
  \caption{
    Architecture of the graph convolution network used to process input scene graphs.
    The input scene graph has $O$ objects and $R$ relationships. Due to weight sharing
    in graph convolutions, the same network can process graphs of any size or topology.
    The notation gconv($D_{in}\to H\to D_{out}$) is graph convolution with input dimension
    $D_{in}$, hidden dimension $H$, and output dimension $D_{out}$.
  }
  \label{tab:gconv-net}
\end{table}

\subsection{Box Regression Network}
We predict bounding boxes for images using a \emph{box regression network}. The input to the
box regression network are the final embedding vectors for objects produced by the graph
convolution network. The output from the box regression network is a predicted bounding box for
the object, parameterized as $(x_0, y_0, x_1, y_1)$ where $x_0, x_1$ are the left and right coordinates
of the box and $y_0,y_1$ are the top and bottom coordinates of the box; all box coordinates are normalized
to be in the range $[0, 1]$. The architecture of the box regression network is shown in
Table~\ref{tab:box-net}.

\begin{table}
  \centering
  \setlength{\tabcolsep}{1mm}
  \scalebox{0.95}{
  \begin{tabular}{|c|c|c|c|}
    \hline
    \textbf{Index} & \textbf{Inputs} & \textbf{Operation} & \textbf{Output Shape} \\
    \hline
    (1) & - & Object embedding vector & 128 \\
    (2) & (1) & Linear($128\to512$) & 512 \\
    (3) & (2) & ReLU & 512 \\
    (4) & (3) & Linear($512\to4$) & 4 \\
    \hline
  \end{tabular}}
  \vspace{1mm}
  \caption{
    Architecture of the box regression network.
  }
  \label{tab:box-net}
\end{table}

\subsection{Mask Regression Network}
We predict segmentation masks for images using a \emph{mask regression network}.
The input to the mask regression network are the final embedding vectors for objects from the
graph convolution network, and the output from the mask regresion network is a $M\times M$
segmentation mask with all elements in the range $(0, 1)$. The mask regression network is composed
of a sequence of upsampling and convolution layers, terminating in a sigmoid nonlinearity; its exact
architecture is shown in Table~\ref{tab:mask-net}.

The main text of the paper states that the mask regression network uses transpose convolution, but
in fact it uses upsampling and stride-1 convolutions as shown in Table~\ref{tab:mask-net}. This error
will be corrected in the camera-ready version of the paper.

\begin{table}
  \centering
  \setlength{\tabcolsep}{1mm}
  \scalebox{0.85}{
  \begin{tabular}{|c|c|c|c|}
    \hline
    \textbf{Index} & \textbf{Inputs} & \textbf{Operation} & \textbf{Output Shape} \\
    \hline
    (1) & - & Object embedding vector & 128 \\
    (2) & (1) & Reshape & $128\times 1 \times 1$ \\
    (3) & (2) & Upsample & $128\times 2 \times 2$ \\
    (4) & (3) & Batch Normalization & $128 \times 2 \times 2$ \\
    (5) & (4) & Conv($3\times3, 128\to 128$) & $128\times 2 \times 2$ \\
    (6) & (5) & ReLU & $128\times2\times2$ \\
    (7) & (6) & Upsample & $128\times 4 \times 4$ \\
    (8) & (7) & Batch Normalization & $128 \times 4 \times 4$ \\
    (9) & (8) & Conv($3\times3, 128\to 128$) & $128\times 4 \times 4$ \\
    (10) & (9) & ReLU & $128\times4\times4$ \\
    (11) & (10) & Upsample & $128\times 8 \times 8$ \\
    (12) & (11) & Batch Normalization & $128 \times 8 \times 8$ \\
    (13) & (12) & Conv($3\times3, 128\to 128$) & $128\times 8 \times 8$ \\
    (14) & (13) & ReLU & $128\times8\times8$ \\
    (15) & (14) & Upsample & $128\times 16 \times 16$ \\
    (16) & (15) & Batch Normalization & $128 \times 16 \times 16$ \\
    (17) & (16) & Conv($3\times3, 128\to 128$) & $128\times 16 \times 16$ \\
    (18) & (17) & ReLU & $128\times16\times16$ \\
    (19) & (18) & conv($1\times1, 128\to1$) & $1\times 16 \times 16$ \\
    (20) & (19) & sigmoid & $1\times16\times16$ \\
    \hline
  \end{tabular}}
  \vspace{1mm}
  \caption{
    Architecture of the mask regression network. For 3D tensors we use $C\times H\times W$
    layout, where $C$ is the number of channels in the feature map and $H$ and $W$ are the
    height and width of the feature map. The notation Conv($K\times K, C_{in}\to C_{out}$)
    is a convolution with $K\times K$ kernels, $C_{in}$ input channels and $C_{out}$ output
    channels; all convolutions are stride 1 with zero padding so that their input and output
    have the same spatial size. Upsample is a $2\times 2$ nearest-neighbor upsampling.
  }
  \label{tab:mask-net}
\end{table}

\subsection{Scene Layout}
The final embedding vectors for objects from the graph convolution network are combined with
the predicted bounding boxes and segmentation masks for objects to give a \emph{scene layout}.
The conversion from vectors, masks, and boxes to scene layouts does not have any learnable parameters.

The scene layout has shape $D\times H\times W$ where $D=128$ is the dimension of embededing vectors
for objects from the graph convolution network and $H\times W=64\times64$ is the output resolution at
which images will be generated.

\subsection{Cascaded Refinement Network}
The scene layout is converted to an image using a \emph{Cascaded Refinement Network} (CRN) consisting of a
number of \emph{Cascaded Refinement Modules} (CRMs).

Each CRM recieves as input the scene layout of shape $D\times H\times W=128\times 64\times 64$ and the
previous feature map, and outputs a new feature map twice the spatial size of the input feature map.
Internally each CRM upsamples the input feature map by a factor of 2, and downsamples the layout using
average pooling the match the size of the upsampled feature map; the two are concatenated and processed
with two convolution layers. A CRM taking input of shape $C_{in}\times H_{in}\times W_{out}$ and producing
an output of shape $C_{out}\times H_{out}\times W_{out}$ (with $H_{out}=2 H_{in}$ and $W_{out}=2 W_{in}$
is denoted as CRM($H_{in}\times W_{in},\; C_{in}\to C_{out}$). The exact architecture of our CRMs is
shown in Table~\ref{tab:crn-module}.

Our Cascaded Refinement Network consists of five Cascaded Refinement Modules. The input to the first
module is Gaussian noise of shape $32\times2\times2$ and the output from the final module is processed
with two final convolution layers to produce the output image. The architecture of the CRN is shown
in Table~\ref{tab:crn-net}.

\begin{table}
  \centering
  \setlength{\tabcolsep}{1mm}
  \scalebox{0.77}{
  \begin{tabular}{|c|c|c|c|}
    \hline
    \textbf{Index} & \textbf{Inputs} & \textbf{Operation} & \textbf{Output Shape} \\
    \hline
    (1) & - & Scene Layout & $D\times H\times W$ \\
    (2) & - & Input features & $C_{in}\times H_{in} \times W_{in}$ \\
    (3) & (1) & Average Pooling & $D\times H_{out} \times W_{out}$ \\
    (4) & (2) & Upsample & $C_{in}\times H_{out} \times W_{out}$ \\
    (5) & (3), (4) & Concatenation & $(D + C_{in})\times H_{out}\times W_{out}$ \\
    (6) & (5) & Conv($3\times3, D+C_{in}\to C_{out}$) & $C_{out}\times H_{out} \times W_{out}$ \\
    (7) & (6) & Batch Normalization & $C_{out}\times H_{out} \times W_{out}$ \\
    (8) & (7) & LeakyReLU & $C_{out}\times H_{out} \times W_{out}$ \\
    (9) & (8) & Conv($3\times 3, C_{out}\to C_{out}$) & $C_{out}\times H_{out} \times W_{out}$ \\
    (10) & (9) & Batch Normalization & $C_{out}\times H_{out} \times W_{out}$ \\
    (11) & (10) & LeakyReLU & $C_{out}\times H_{out} \times W_{out}$ \\
    \hline
  \end{tabular}}
  \vspace{1mm}
  \caption{
    Architecture of a Cascaded Refinement Module
    CRM($H_{in}\times W_{in},\;C_{in}\to C_{out})$. The module accepts as input
    the scene layout, and an input feature map of shape
    $C_{in}\times H_{in}\times W_{in}$ and produces as output a feature map of
    shape $C_{out}\times H_{out}\times W_{out}$ where $H_{out}=2 H_{in}$ and
    $W_{out} = 2 W_{in}$. For LeakyReLU nonlinearites we use negative slope 0.2.
  }
  \label{tab:crn-module}
\end{table}

\begin{table}
  \centering
  \setlength{\tabcolsep}{1mm}
  \scalebox{0.92}{
  \begin{tabular}{|c|c|c|c|}
    \hline
    \textbf{Index} & \textbf{Inputs} & \textbf{Operation} & \textbf{Output Shape} \\
    \hline
    (1) & - & Scene Layout & $128 \times 64 \times 64$ \\
    (2) & - & Gaussian Noise & $32\times 2 \times 2$ \\
    (3) & (1), (2) & CRN($2\times2,\; 32\to1024$) & $1024\times4\times4$ \\
    (4) & (1), (3) & CRN($4\times4,\; 1024\to512$) & $512\times8\times8$ \\
    (5) & (1), (4) & CRN($8\times4,\; 512\to256$) & $256\times16\times16$ \\
    (6) & (1), (5) & CRN($16\times16,\; 256\to128$) & $128\times32\times32$ \\
    (7) & (1), (6) & CRN($32\times32,\; 128\to64$) & $64\times64\times64$ \\
    (8) & (7) & Conv($3\times3, 64\to64$) & $64\times64\times64$ \\
    (9) & (8) & LeakyReLU & $64\times64\times64$ \\
    (10) & (9) & Conv($1\times1, 64\to3$) & $3\times64\times64$ \\
    \hline
  \end{tabular}}
  \vspace{1mm}
  \caption{
    Architecture of our Cascaded Refinement Network. CRM is a Cascaded Refinement Module,
    shown in Table~\ref{tab:crn-module}. LeakyReLU uses a negative slope of 0.2.
  }
  \label{tab:crn-net}
\end{table}

\subsection{Batch Normalization in the Generator}
Most implementations of batch normalization operate in two modes. In \emph{train} mode,
minibatches are normalized using the empirical mean and variance of features; in \emph{eval}
mode a running mean of feature means and variances are used to normalize minibatches instead.
We found that training models in \emph{train} mode and running them in \emph{eval} mode at
test-time led to significant image artifacts. To overcome this limitation while still
benefitting from the optimization benefits that batch normalization provides, we train
our models for 100K iterations using batch normalization in \emph{train} mode, then
continue training for an additional 900K iterations with batch normalization in \emph{eval}
mode.

Since discriminators are not used at test-time, batch normalization in the discriminators
is always used in \emph{train} mode.

\subsection{Object Discriminator}
Our object discriminator $D_{obj}$ inputs image pixels corresponding to objects in
real or generated images; objects are cropped using their bounding boxes to a spatial
size of $32\times32$ using differentiable bilinear interpolation. The object discriminator
serves two roles: it classifies objects as real or fake, and also uses an \emph{auxiliary classifier}
which attempts to classify each object. The exact architecture of our object discriminator
is shown in Table~\ref{tab:discriminator-obj}.

\begin{table}
  \centering
  \setlength{\tabcolsep}{1mm}
  \scalebox{0.92}{
  \begin{tabular}{|c|c|c|c|}
    \hline
    \textbf{Index} & \textbf{Inputs} & \textbf{Operation} & \textbf{Output Shape} \\
    (1) & - & Object crop & $3\times32\times32$ \\
    (2) & (1) & Conv($4\times4, 3\to64, s2$) & $64\times16\times16$ \\
    (3) & (2) & Batch Normalization & $64\times16\times16$ \\
    (4) & (3) & LeakyReLU & $64\times16\times16$ \\
    (5) & (4) & Conv($4\times4, 64\to128, s2$) & $128\times8\times8$ \\
    (6) & (5) & Batch Normalization & $128\times32\times32$ \\
    (7) & (6) & LeakyReLU & $128\times32\times32$ \\
    (8) & (7) & Conv($4\times4, 128\to256, s2$) & $256\times4\times4$ \\
    (9) & (8) & Global Average Pooling & $256$ \\
    (10) & (9) & Linear($256\to1024$) & $1024$ \\
    (11) & (10) & Linear($1024\to1$) & 1 \\
    (12) & (10) & Linear($1024\to|\mathcal{C}|$) & $|\mathcal{C}|$ \\
    \hline
  \end{tabular}}
  \vspace{1mm}
  \caption{
    Architecture of our object discriminator $D_{obj}$. The input to the object discriminator
    is a $32\times32$ crop of an object in either a generated or real image. The object
    discriminator outputs both a score for real / fake (11) and a classification score over the
    object categories $\mathcal{C}$ (12). In this model all convolution layers have stride 2 and
    no zero padding. LeakyReLU uses a negative slope of 0.2.
  }
  \label{tab:discriminator-obj}
\end{table}

\subsection{Image Discriminator}
Our image discriminator $D_{img}$ inputs a real or fake image, and classifies an
overlapping grid of $8\times8$ image patches from its input image as real or fake.
The exact architecture of our image discriminator is shown in Table~\ref{tab:discriminator-img}.

\subsection{Higher Image Resolutions}
We performed preliminary experiments with a version of our model that produces $128\times128$
images rather than $64\times64$ images. For these models we compute the scene layout at $128\times128$
rather than at $64\times64$; we also add an extra Cascaded Refinement Module to our Cascaded Refinement
Network; we add one additional convolutional layer to both $D_{obj}$ and $D_{img}$, and for these
models $D_{obj}$ receives a $64\times64$ crop of objects rather than a $32\times32$ crop. During trainging
we reduce the batch size from 32 to 24.

The images in Figure 6 from the main paper were generated from a version of our model trained to
produce $128\times128$ images from Visual Genome.

\begin{table}
  \centering
  \setlength{\tabcolsep}{1mm}
  \scalebox{0.92}{
  \begin{tabular}{|c|c|c|c|}
    \hline
    \textbf{Index} & \textbf{Inputs} & \textbf{Operation} & \textbf{Output Shape} \\
    (1) & - & Image & $3\times64\times64$ \\
    (2) & (1) & Conv($4\times4,3\to64, s2$) & $64\times32\times32$ \\
    (3) & (2) & Batch Normalization & $64\times32\times32$ \\
    (4) & (3) & LeakyReLU & $64\times32\times32$ \\

    (5) & (4) & Conv($4\times4,64\to128, s2$) & $128\times16\times16$ \\
    (6) & (5) & Batch Normalization & $128\times16\times16$ \\
    (7) & (6) & LeakyReLU & $128\times16\times16$ \\
    (8) & (7) & Conv($4\times4,128\to256, s2$) & $256\times8\times8$ \\
    (9) & (8) & Conv($1\times1,256\to1$) & $1\times8\times8$ \\
    \hline
  \end{tabular}}
  \vspace{1mm}
  \caption{
    Architecture of our image discriminator $D_{img}$. The input to the image discriminator
    is either a real or fake image, and it classifies an overlapping $8\times8$ grid of
    patches in the input image as either real or fake. All but the final convolution have
    a stride of 2, and all convolutions use no padding. LeakyReLU uses a negative slope of 0.2.
  }
  \label{tab:discriminator-img}
\end{table}

\section{Image Loss Functions}
\begin{figure*}
  \newlength{\lossimgsize}
  \setlength{\lossimgsize}{0.138\textwidth}
  \centering
  \begin{minipage}{\lossimgsize}
    \centering
    \textbf{GT Image}
  \end{minipage}
  \begin{minipage}{\lossimgsize}
    \centering
    \textbf{Scene \\ Graph}
  \end{minipage}
  \begin{minipage}{\lossimgsize}
    \centering
    \textbf{Ours \\ (L1 only)}
  \end{minipage}
  \begin{minipage}{\lossimgsize}
    \centering
    \textbf{Ours \\ (No  $D_{obj}$ )}
  \end{minipage}
  \begin{minipage}{\lossimgsize}
    \centering
    \textbf{Ours \\ (No } $D_{img}$ \textbf{)}
  \end{minipage}
  \begin{minipage}{\lossimgsize}
    \centering
    \textbf{Ours \\ (full model)}
  \end{minipage}
  \begin{minipage}{\lossimgsize}
    \centering
    \textbf{Ours \\ (GT Layout)}
  \end{minipage} \\
  \includegraphics[width=\lossimgsize]{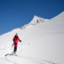}
  \includegraphics[width=\lossimgsize]{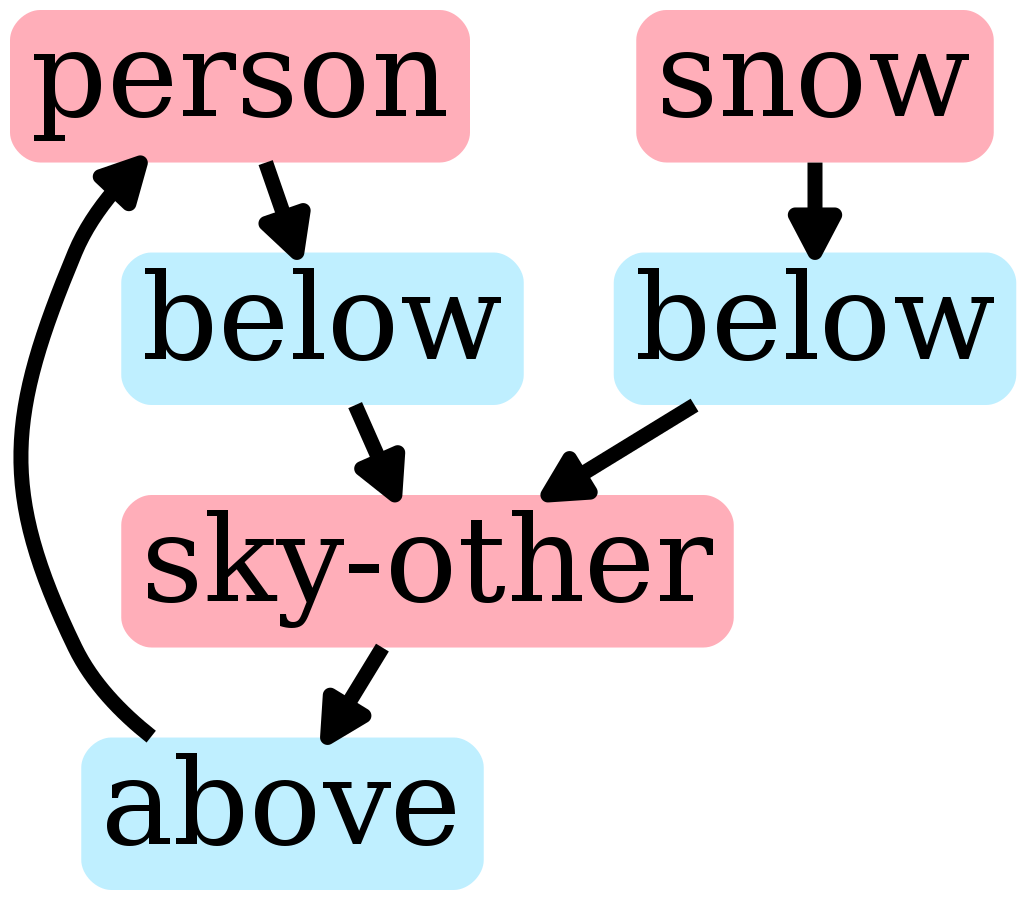}
  \includegraphics[width=\lossimgsize]{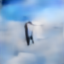}
  \includegraphics[width=\lossimgsize]{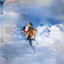}
  \includegraphics[width=\lossimgsize]{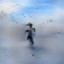}
  \includegraphics[width=\lossimgsize]{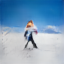}
  \includegraphics[width=\lossimgsize]{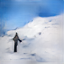} \\
  \includegraphics[width=\lossimgsize]{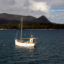}
  \hspace*{6.6mm}\includegraphics[height=\lossimgsize]{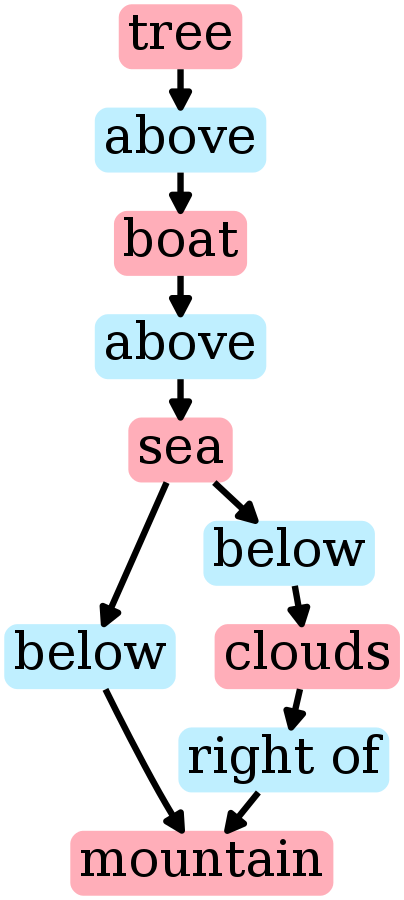}\hspace*{6.6mm}
  \includegraphics[width=\lossimgsize]{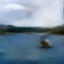}
  \includegraphics[width=\lossimgsize]{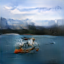}
  \includegraphics[width=\lossimgsize]{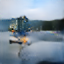}
  \includegraphics[width=\lossimgsize]{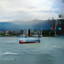}
  \includegraphics[width=\lossimgsize]{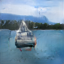} \\
  \includegraphics[width=\lossimgsize]{figures/coco_64_supp/gt_000617.png}
  \includegraphics[width=\lossimgsize]{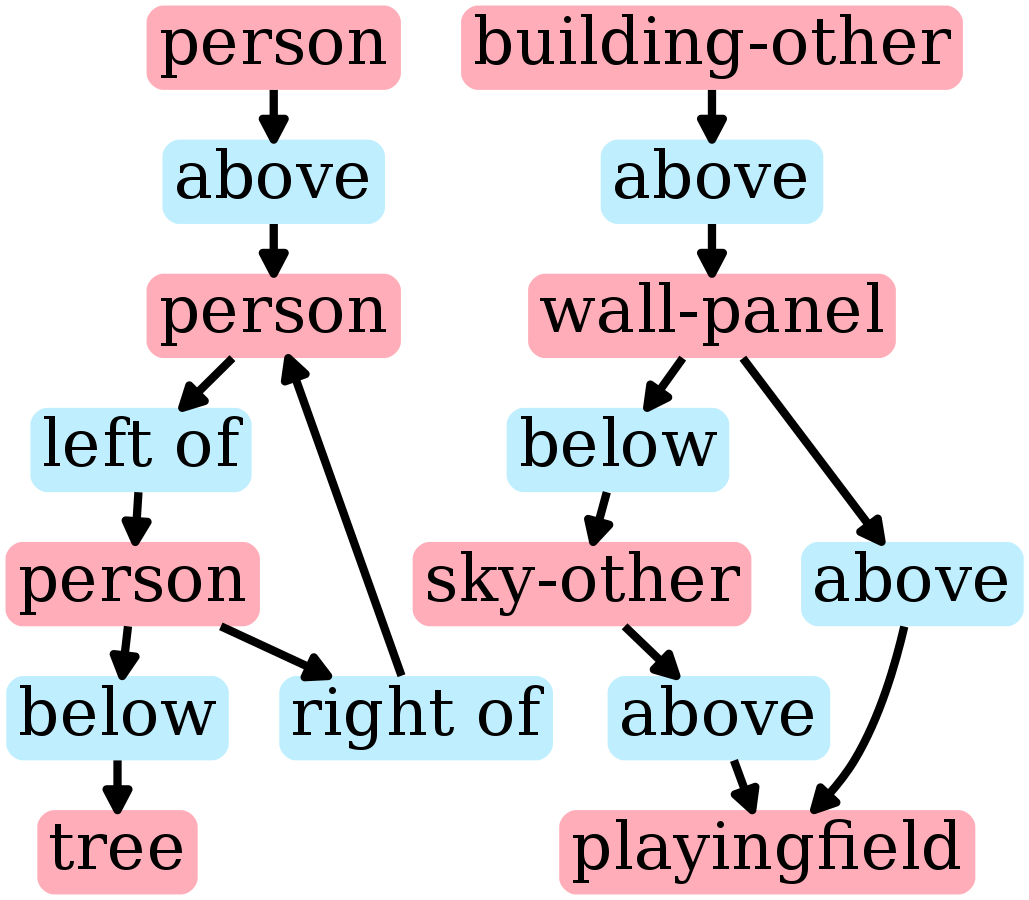}
  \includegraphics[width=\lossimgsize]{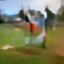}
  \includegraphics[width=\lossimgsize]{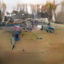}
  \includegraphics[width=\lossimgsize]{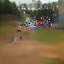}
  \includegraphics[width=\lossimgsize]{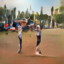}
  \includegraphics[width=\lossimgsize]{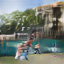} \\
  \includegraphics[width=\lossimgsize]{figures/coco_64_supp/gt_000835.png}
  \hspace*{1.8mm}\includegraphics[height=\lossimgsize]{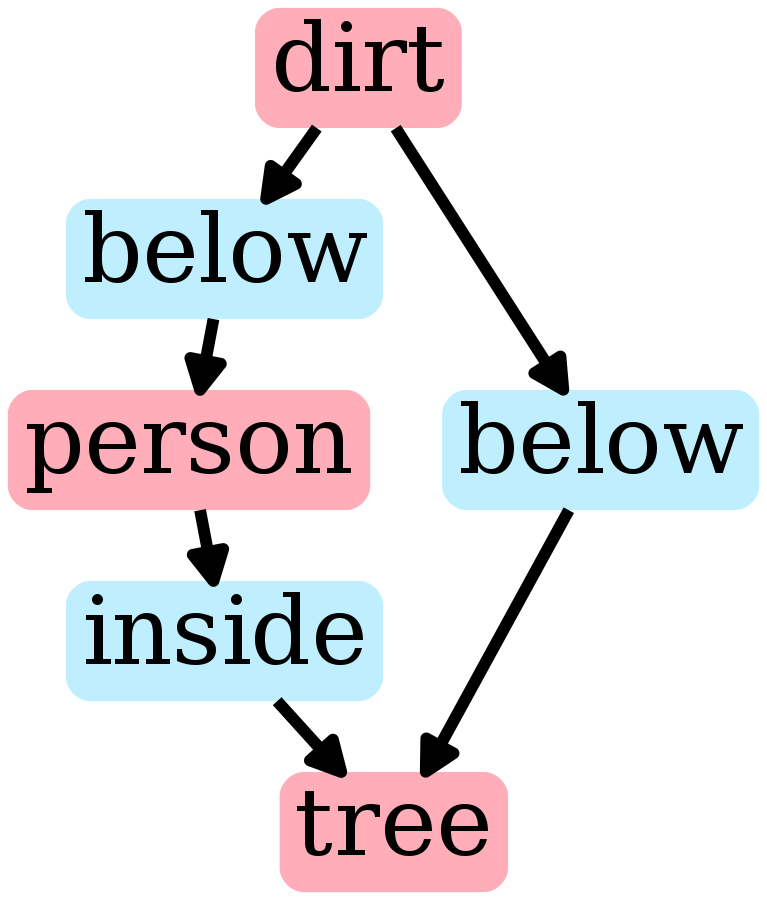}\hspace*{1.8mm}
  \includegraphics[width=\lossimgsize]{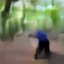}
  \includegraphics[width=\lossimgsize]{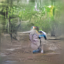}
  \includegraphics[width=\lossimgsize]{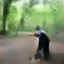}
  \includegraphics[width=\lossimgsize]{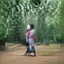}
  \includegraphics[width=\lossimgsize]{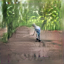} \\
  \includegraphics[width=\lossimgsize]{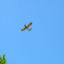}
  \hspace*{5.2mm}\includegraphics[height=\lossimgsize]{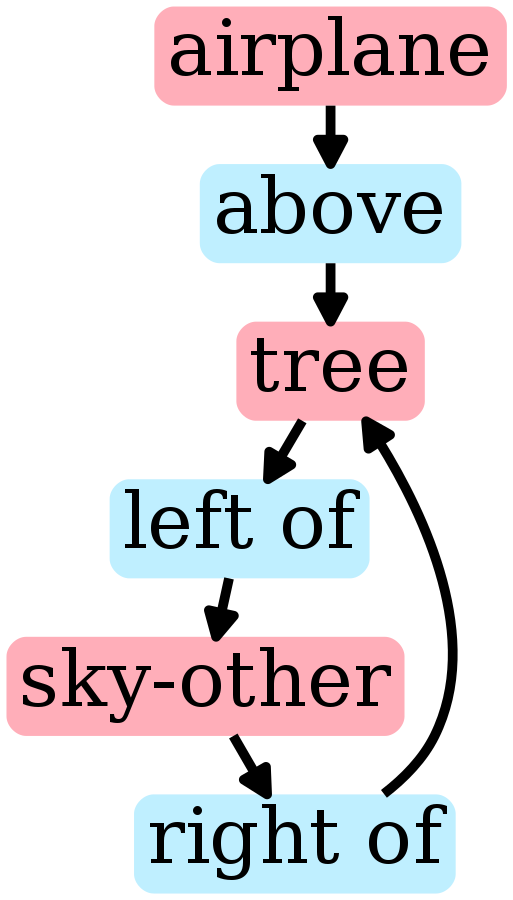}\hspace*{5.2mm}
  \includegraphics[width=\lossimgsize]{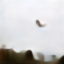}
  \includegraphics[width=\lossimgsize]{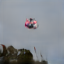}
  \includegraphics[width=\lossimgsize]{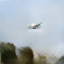}
  \includegraphics[width=\lossimgsize]{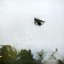}
  \includegraphics[width=\lossimgsize]{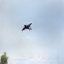} \\
  \includegraphics[width=\lossimgsize]{figures/coco_64_supp/gt_000959.png}
  \includegraphics[width=\lossimgsize]{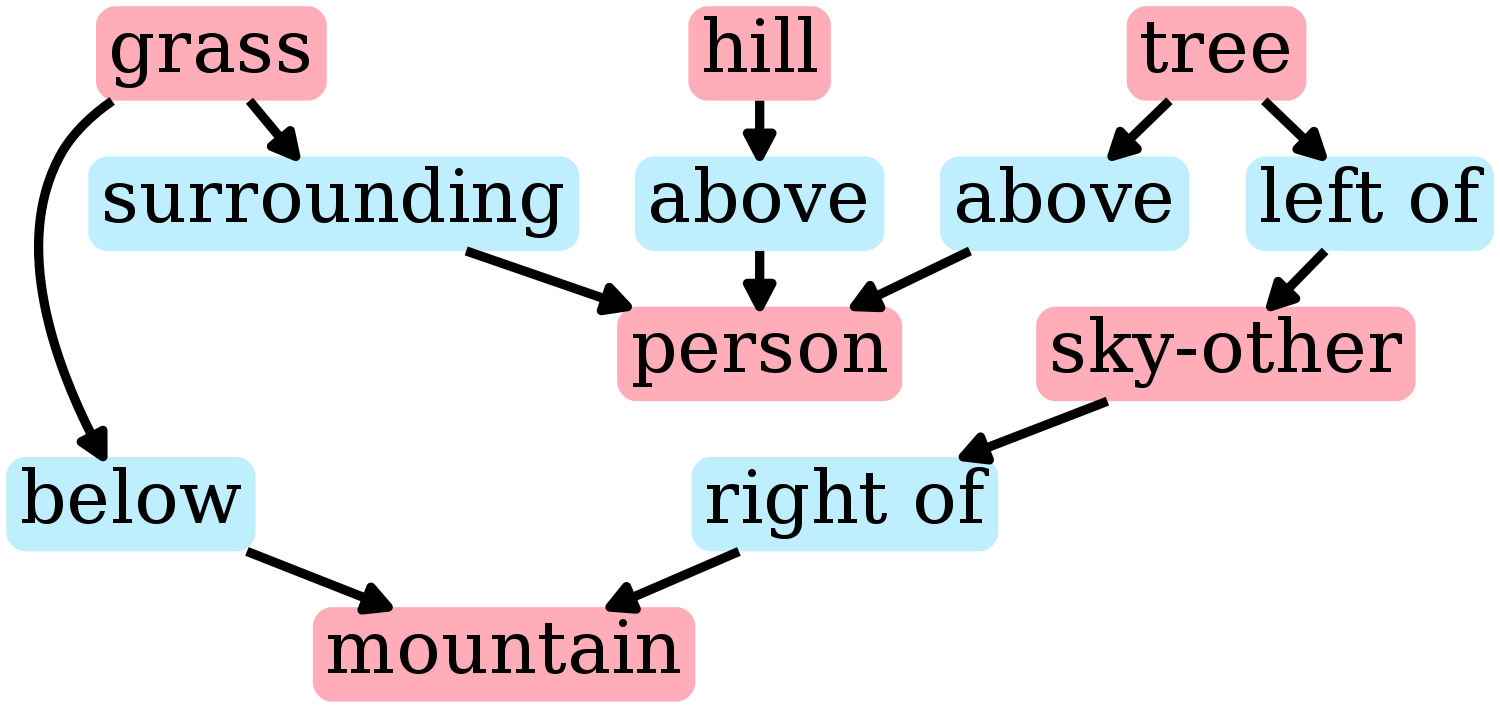}
  \includegraphics[width=\lossimgsize]{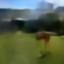}
  \includegraphics[width=\lossimgsize]{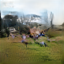}
  \includegraphics[width=\lossimgsize]{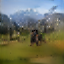}
  \includegraphics[width=\lossimgsize]{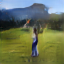}
  \includegraphics[width=\lossimgsize]{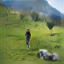} \\
  \includegraphics[width=\lossimgsize]{figures/coco_64_supp/gt_000980.png}
  \includegraphics[width=\lossimgsize]{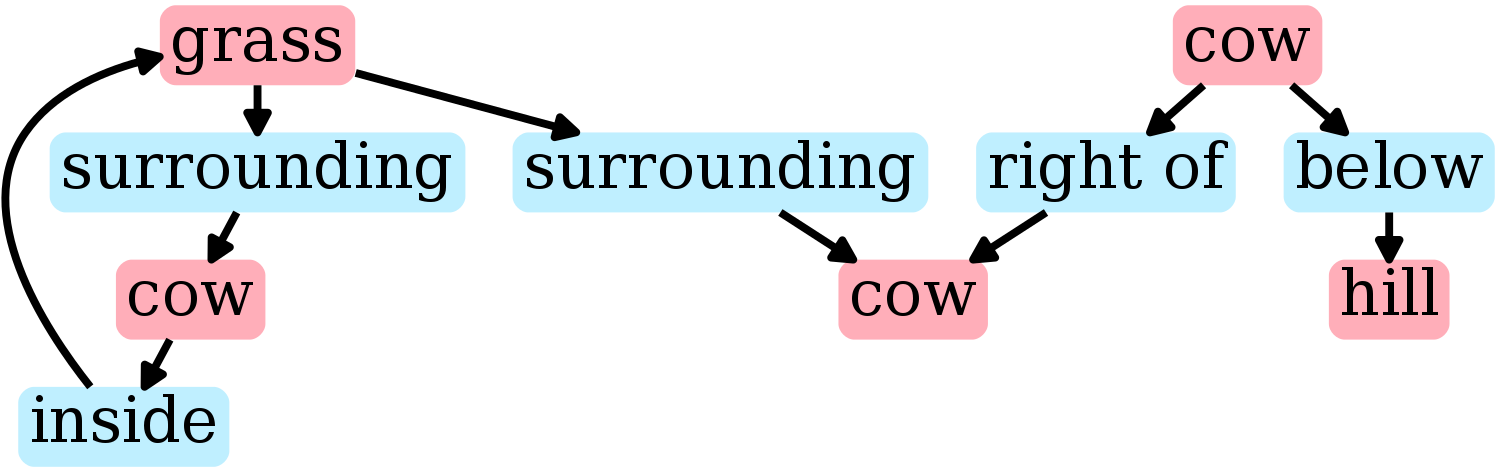}
  \includegraphics[width=\lossimgsize]{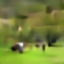}
  \includegraphics[width=\lossimgsize]{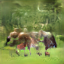}
  \includegraphics[width=\lossimgsize]{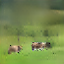}
  \includegraphics[width=\lossimgsize]{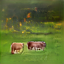}
  \includegraphics[width=\lossimgsize]{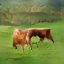} \\*[-2mm]
  \rule{\textwidth}{1pt} \\*[1mm]
  \includegraphics[width=\lossimgsize]{figures/coco_64_supp/gt_001029.png}
  \hspace*{1mm}\includegraphics[height=\lossimgsize]{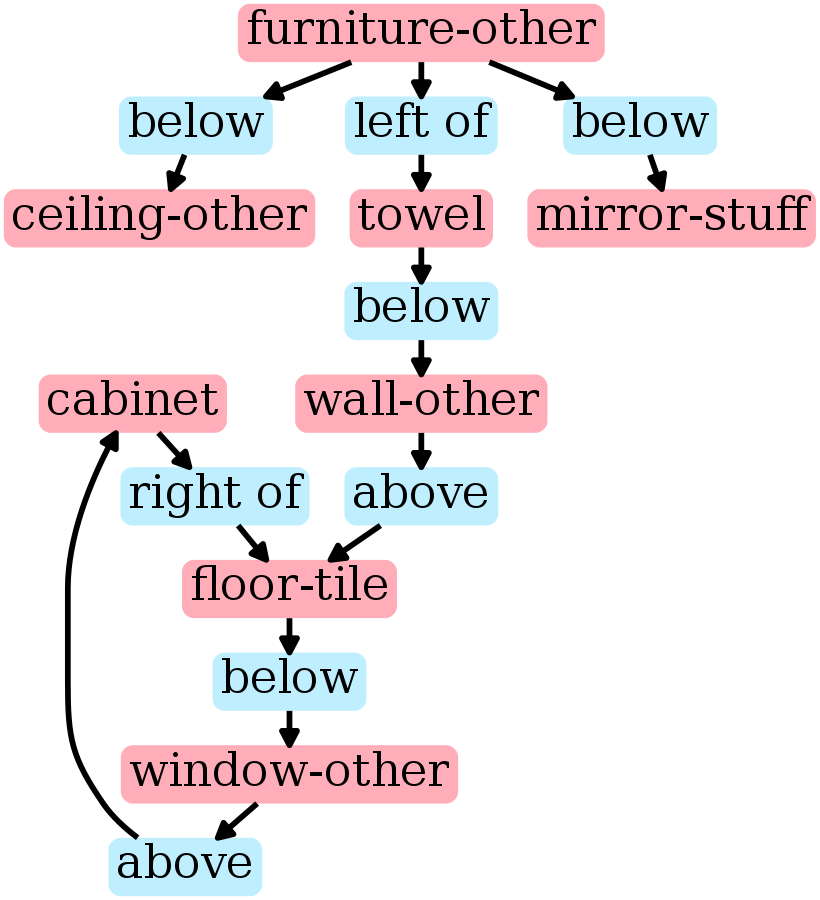}\hspace*{1mm}
  \includegraphics[width=\lossimgsize]{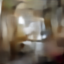}
  \includegraphics[width=\lossimgsize]{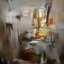}
  \includegraphics[width=\lossimgsize]{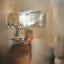}
  \includegraphics[width=\lossimgsize]{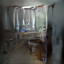}
  \includegraphics[width=\lossimgsize]{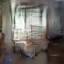} \\
  \caption{
    Example images generated from our model and ablations on COCO. We show the original image,
    the synthetic scene graph generated from the COCO annotations for the image, and results from
    several versions of our model. Our model with no discriminators (L1 only) tends to be overly smooth;
    omitting the object discriminator (No $D_{obj}$) causes objects to be less recognizable; omitting
    the image discriminator (No $D_{img}$) leads to low-level image artifacts. Using the ground-truth
    rather than predicted layout (GT Layout) tends to result in higher quality images. The bottom
    row shows a typical failure case, where all versions of our model struggle with complex scene
    graphs for indoor scenes. Graphs best viewed with magnification.
  }
  \label{fig:more-qual}
\end{figure*}

In Figure~\ref{fig:more-qual} we show additional qualitative results from our model trained
on COCO, comparing the results from different ablated versions of our model.

Omitting the discriminators from the model (L1 only) tends to produce images that are overly
smoothed. Without the object discriminator (No $D_{obj}$) objects tend to be less recognizable,
and without the image discriminator (No $D_{img}$) the generated images tend to appear less
realistic overall, with low-level artifacts. Our model trained to use ground-truth layouts
rather than predicting its own layouts (GT Layout) tends to produce higher-quality images,
but requires both bounding-box and segmentation mask annotations at test-time.

The bottom row of Figure~\ref{fig:more-qual} also shows a typical failure case,
where all models struggle to synthesize a realistic image from a complex scene graph for an
indoor scene.

\begin{figure*}
  \centering
  \includegraphics[width=0.48\textwidth]{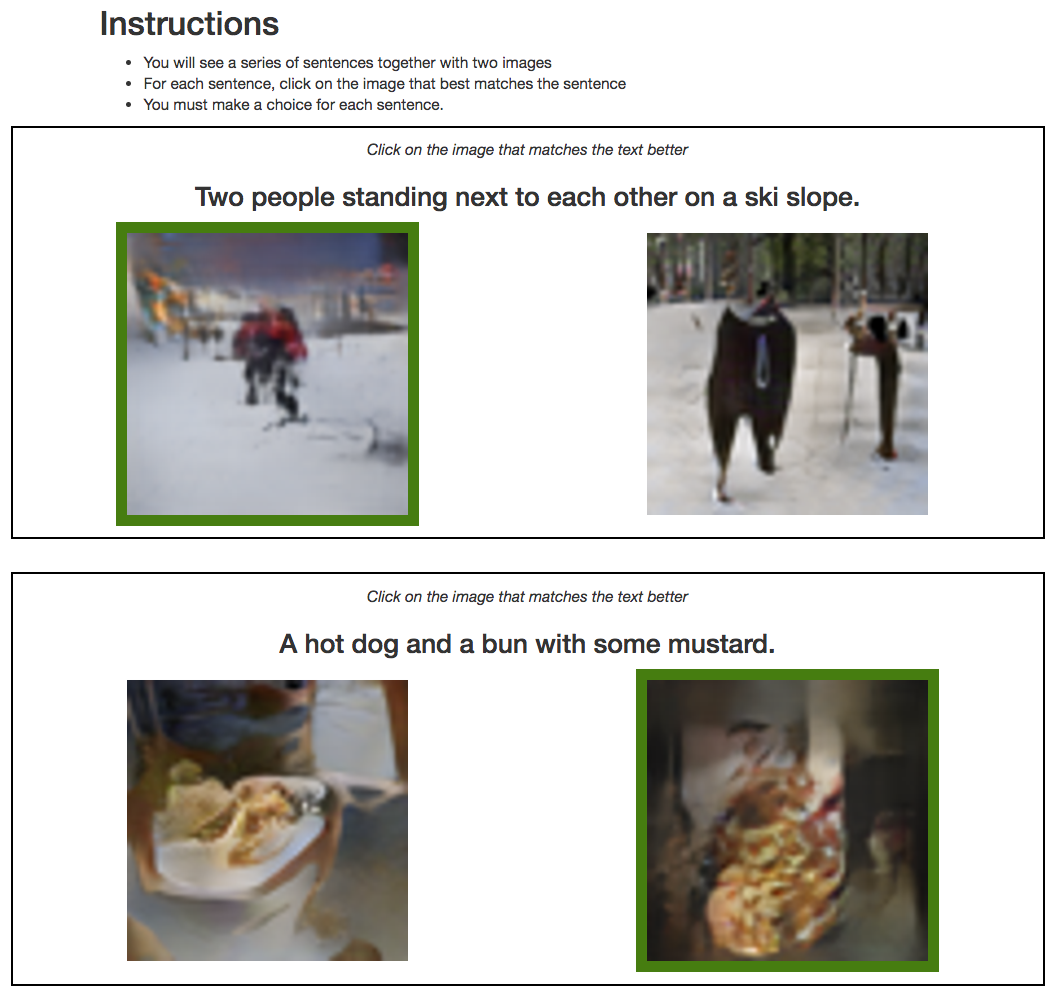}
  \raisebox{18mm}{\includegraphics[width=0.48\textwidth]{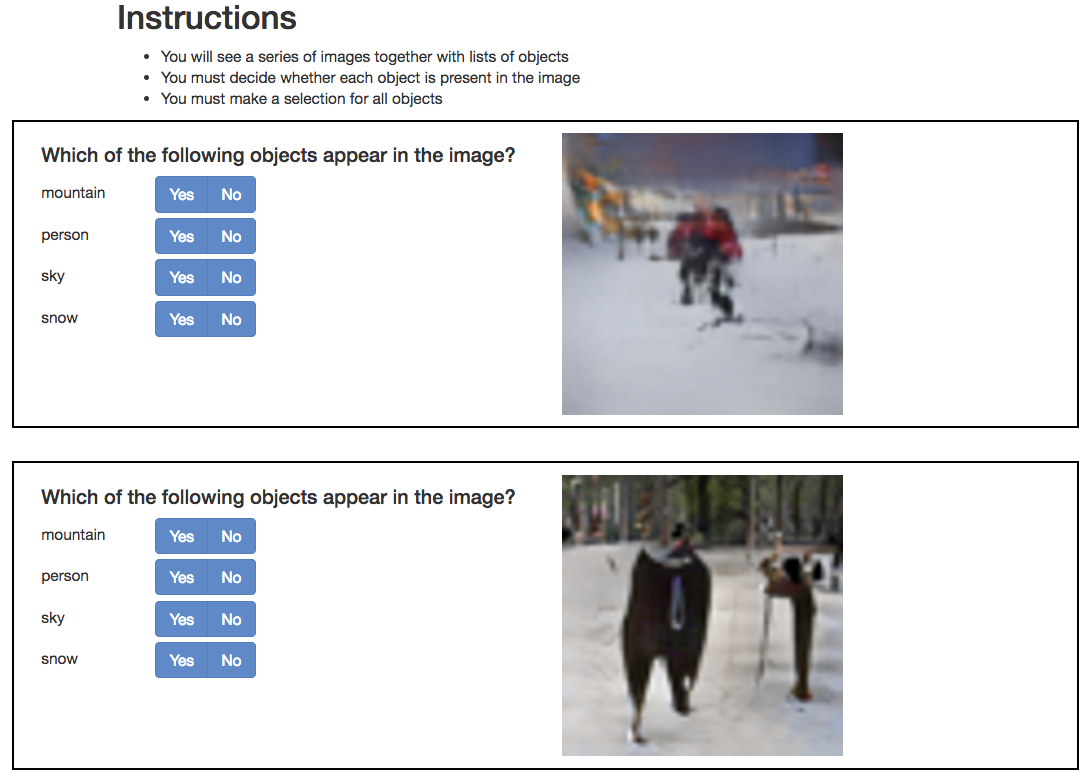}}
  \caption{
    Screenshots of the user interfaces for our user studies on Amazon Mechanical Turk.
    \textbf{Left:} User interface for the user study from Figure 7 of the main paper.
    We show users an image generated by StackGAN from a COCO caption, and an image 
    generated with our method from a scene graph built from the COCO object annotations
    corresponding to the caption. We ask users to select the image that best matches
    the caption.
    \textbf{Right:} User interface for the user study from Figure 8 of the main paper.
    We show again show users images generated using StackGAN and our method, and we
    ask users which COCO objects are present in each image.
  }
  \label{fig:user-study-ui}
\end{figure*}

\section{User Study}
As discussed in Section 4.5 of the main paper, we perform two user studies on Amazon
Mechanical Turk to compare the perceptual quality of images generated from our method
with those generated using StackGAN.

In the first user study, we show users an image generated from a COCO caption using
StackGAN, and an image generated using our method from a scene graph built from the
COCO object annotations corresponding to the caption. We ask users to select the image that
better matches the caption. In each trial of this user study the order of our image and the
image from StackGAN are randomized.

In the second user study, we again show users images generated using both methods,
and we ask users to select the COCO objects that are visible in the image.
In this experiment, if a single image contains multiple instances of the same object
category then we only ask about its presence once. In each Mechanical Turk HIT users
see an equal number of results from StackGAN and our method, and the order in which
they are presented is randomized.

For both studies we use 1024 images from each method generated from COCO val annotations.
All images are seen by three workers, and we report all results using majority opinions.

StackGAN produces $256\times256$ images, but our method produces $64\times64$ images.
To prevent the differing image resolution from affecting worker opinion, we downsample
StackGAN results to $64\times64$ using bicubic interpolation before presenting them to
users.

\clearpage

\end{document}